\documentclass[letterpaper, 10 pt, conference]{ieeeconf}  % Comment this line out
\IEEEoverridecommandlockouts                              % This command is only
\usepackage[english]{babel}
\usepackage[T1]{fontenc}

\usepackage[nolist]{acronym}
\usepackage[cmex10]{amsmath}
\usepackage{amssymb}
\usepackage{bm}
\usepackage{booktabs}
\usepackage{color}
\usepackage{float}
\usepackage{glossaries}  % for acronyms
\usepackage{graphicx}
\usepackage{hyperref}
\usepackage{cite}

\usepackage{cleveref}
\usepackage[font={small}]{caption}
\usepackage{import}
\usepackage{mathtools}
\usepackage{multirow}
\usepackage{outline}
\usepackage{paralist}
\usepackage{siunitx}
\usepackage{stmaryrd}
\usepackage{systeme}
\usepackage{url}
\usepackage{tikz}
\usepackage{booktabs}
\usetikzlibrary{tikzmark}
\usepackage{subfig}
\usepackage{soul}

\let\labelindent\relax
\usepackage{enumitem}       % Style for enum and itemize 
\usepackage{pifont}         % More styles for bullets

\setlist{leftmargin=5.5mm}  % Reduce left margin.

\usetikzlibrary{positioning}
\usetikzlibrary{calc}

\usepackage{mathtools}
\usepackage{xparse}
\usepackage{amssymb}
\usepackage{bm}
\usepackage{xfrac}
\usepackage{tensor}
\usepackage{eqparbox}
\usepackage{xcolor}
\usepackage{pifont}
\usepackage[all,cmtip]{xy}
\DeclarePairedDelimiterX{\norm}[1]{\lVert}{\rVert}{#1}
\newcommand{\stforall}{\;|\; \forall \; }

\newcommand{\G}[1][]{\mathbb{G}_{\scalebox{0.6}{$#1$}}}       % Symmetry Group 
            
\newcommand{\CyclicGroup}[1][]{\mathbb{C}_{#1}}        % Cyclic Group 
\newcommand{\GLGroup}{\mathbb{GL}}          % General linear Group 
\newcommand{\g}{g}       % Group member 
\newcommand{\Glact}{\mathrel{\scalebox{0.8}{$\triangleright$}}}             % Left group action    g \Glact x
\newcommand{\Gcomp}{\mathrel{\scalebox{0.8}{$\circ$}}}                      % Group binary/composition operator 

\newcommand{\rep}[1][]{  % Representation \rep[<Vector Space Symbol>]{<group element symbol>}
    \rho_{\hbox{\scalebox{0.6}{$#1$}}}
} 

\newcommand{\irrep}[2][]{
    {\bar{\rho}_{\scriptscriptstyle{#1}}
    \def\temp{#2}\ifx\temp\empty
    \else
      (#2)%
    \fi
    }
}

\newcommand{\homomorphismDiag}[5]{
    \xymatrix{
        #1 \ar@{-}[r]^{#3}    \ar[d]^{#5}    & #1 \ar[d]^{#5} \\
        #2 \ar@{-}[r]^{#4}                   & #2
    }
}

\newcommand{\isomorphismDiag}[5]{
    \xymatrix{
        #1 \ar@{-}[r]^{#3}    \ar@{-}[d]^{#5} & #1 \ar@{-}[d]^{#5} \\
        #2 \ar@{-}[r]^{#4}                   & #2
    }
}

\newcommand{\identity}[1][]{\bm{1}}

\newcommand{\measure}[1][]{\mu_{#1}}     % Probability measure

\newcommand{\equivObsSpaceDual}[1][_{\measure[t]}^{*\G}]
{
    \tensor*[]{\scalebox{0.8}{$\mathcal{X}$}}{#1}
}
\newtheorem{theorem}{Theorem}

\newtheorem{lemma}{Lemma}

\newtheorem{definition}{Definition}

\newtheorem{proposition}{Proposition}

\newcommand{\ubcolor}[3][awesomeblue]{{
        \color{#1}{
            \underbrace{\color{black}{#2}}_{#3}
        }
    }}

\usepackage{makecell}

\usepackage{amsmath,amsfonts,bm}

\def\eqref#1{equation~\ref{#1}}
\def\plaineqref#1{(\ref{#1})}
\def\1{\bm{1}}

\def\vaa{{\bm{a}}}

\def\vs{{\bm{s}}}

\def\vx{{\bm{x}}}

\def\vsA{{\mathcal{A}}}

\def\vsS{{\mathcal{S}}}

\def\vsX{{\mathcal{X}}}
\def\vsY{{\mathcal{Y}}}

\DeclareMathAlphabet{\mathsfit}{\encodingdefault}{\sfdefault}{m}{sl}
\SetMathAlphabet{\mathsfit}{bold}{\encodingdefault}{\sfdefault}{bx}{n}

\newcommand{\R}{\mathbb{R}}

\definecolor{gray}{rgb}{0.6, 0.7, 0.7}
\definecolor{awesomeblue}{rgb}{0.054, 0.415, 0.505}
\definecolor{awesomeorange}{rgb}{0.570, 0.458, 0.0912}
\definecolor{my_blue}{rgb}{0, 0.4470, 0.7410}
\definecolor{my_yellow}{rgb}{0.9290, 0.6940, 0.1250}
\definecolor{my_purple}{rgb}{0.4940, 0.1840, 0.5560}
\definecolor{my_green}{rgb}{0.4660, 0.6740, 0.1880}
\definecolor{my_red}{rgb}{0.6350, 0.0780, 0.1840}

\newacronym{mdp}{MDP}{Markov decision process}
\newacronym{dof}{DoF}{degrees of freedom}
\newacronym{cot}{CoT}{Cost of Transport}
\newacronym{RSI}{RSI}{Symmetry Index}
\newacronym{ood}{OOD}{out-of-distribution}
\newacronym{PPO}{PPO}{Proximal Policy Optimization}
\newacronym{PPOaug}{PPOaug}{\gls{PPO} with data-augmentation}
\newacronym{PPOsymm}{PPOeqic}{PPO with hard equivariance / invariance symmetry constraints}
\glsunset{PPOaug}
\glsunset{PPOsymm}
\title{\LARGE \bf Leveraging Symmetry in RL-based Legged Locomotion Control}

\author{
    Zhi Su$^{*, 2}$, 
    Xiaoyu Huang$^{*, 1}$, 
    Daniel Ordoñez-Apraez$^{3}$, 
    Yunfei Li$^{2}$, 
    Zhongyu Li$^{1}$, 
    Qiayuan Liao$^{1}$, \\
    Giulio Turrisi$^{3}$, 
    Massimiliano Pontil$^{3}$, 
    Claudio Semini$^{3}$, 
    Yi Wu$^{2,4}$, 
    Koushil Sreenath$^{1}$%
    \thanks{*Equal contribution. $^{1}$UC Berkeley, CA, USA. $^{2}$Institute for Interdisciplinary Information Sciences, Tsinghua University, Beijing, China. $^{3}$Istituto Italiano di Tecnologia, Italy. $^{4}$Shanghai Qi Zhi Institute, Shanghai, China. Contact: haytham.huang@berkeley.edu}%
}
\usepackage{eso-pic}
\AddToShipoutPictureBG*{%
  \AtPageUpperLeft{%
    \setlength{\unitlength}{1mm}%
    \put(-9.5,-9){\makebox(\paperwidth,0)[c]{\parbox{0.9\textwidth}{2024 IEEE/RSJ International Conference on Intelligent Robots and Systems (IROS), pp. 6899-6906}}}
  }
}

\AddToShipoutPictureBG*{%
  \AtPageUpperLeft{%
    \setlength{\unitlength}{1mm}%
    \put(-9.5,-14){\makebox(\paperwidth,0)[c]{\parbox{0.9\textwidth}{DOI: \href{https://ieeexplore.ieee.org/document/10802439}{10.1109/IROS58592.2024.10802439}}}}
  }
}

\begin{document}

\maketitle

\begin{abstract}
    % Deep 
    Model-free reinforcement learning is a promising approach for autonomously solving challenging robotics control problems, but faces exploration difficulty without information about the robot's morphology. The under-exploration of multiple modalities with symmetric states leads to behaviors that are often unnatural and sub-optimal. This issue becomes particularly pronounced in the context of robotic systems with morphological symmetries, such as legged robots for which the resulting asymmetric and aperiodic behaviors compromise performance, robustness, and transferability to real hardware. To mitigate this challenge, we can leverage symmetry to guide and improve the exploration in policy learning via equivariance / invariance constraints. We investigate the efficacy of two approaches to incorporate symmetry: modifying the network architectures to be strictly equivariant / invariant, and leveraging data augmentation to approximate equivariant / invariant actor-critics. We implement the methods on challenging loco-manipulation and bipedal locomotion tasks and compare with an unconstrained baseline. We find that the strictly equivariant policy consistently outperforms other methods in sample efficiency and task performance in simulation. Additionaly, symmetry-incorporated approaches exhibit better gait quality, higher robustness and can be deployed zero-shot to hardware.
\end{abstract}

\section{Introduction} 

The field of robotics has witnessed a surge in the adoption of data-driven reinforcement learning (RL) methods to tackle the control problems of legged locomotion \cite{peng2020learning, li2024reinforcement}, navigation \cite{wijmans2019dd}, and manipulation \cite{singh2019end}. This trend is primarily fueled by the ability of these methods to (i) cope with phenomena that impact the system evolution but are challenging to model analytically, (ii) 
autonomously acquire control strategies without the need of extensive domain knowledge. 
However, commonly-used model-free RL methods often treat the robot as a black-box system; by neglecting analytical models of dynamics, they often remain agnostic to the properties of robot's morphology.
Furthermore, these methods face exploration difficulties to learn multi-modalities, especially symmetric modalities\cite{lee2019efficient} where the under-exploration of some modes leads to asymmetric behaviors that are often unnatural and sub-optimal. For example, failure to fully capture the two symmetric modalities of bipedal locomotion leads to limping behaviors and reduced control performance, compromising robustness and transferability to real hardware.

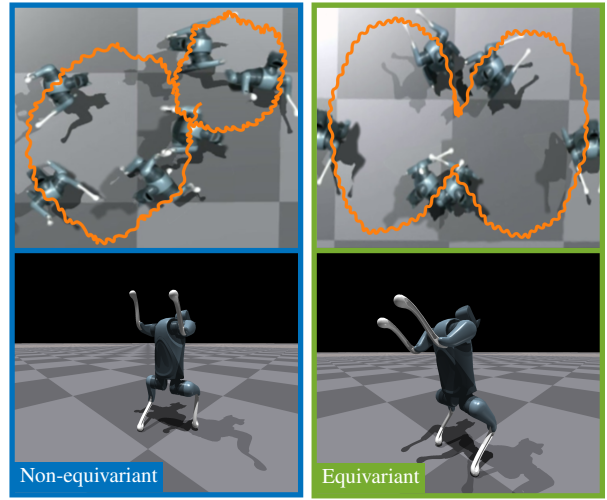
\begin{figure}[t!]
    \centering
    \begin{tikzpicture}[inner sep=0pt,line width=2pt]
      \node (mlpplot) at (0,0) {\includegraphics[width=0.37\linewidth,angle=90, trim={10 0 10 0}, clip]{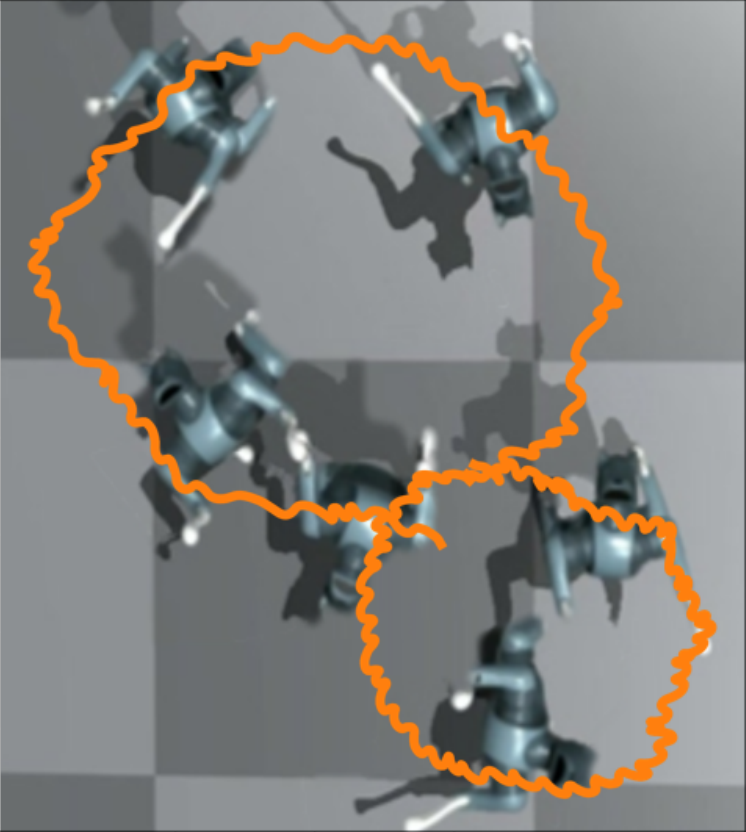}};
      \node[right=5pt of mlpplot] (augplot) {\includegraphics[width=0.37\linewidth,angle=90]{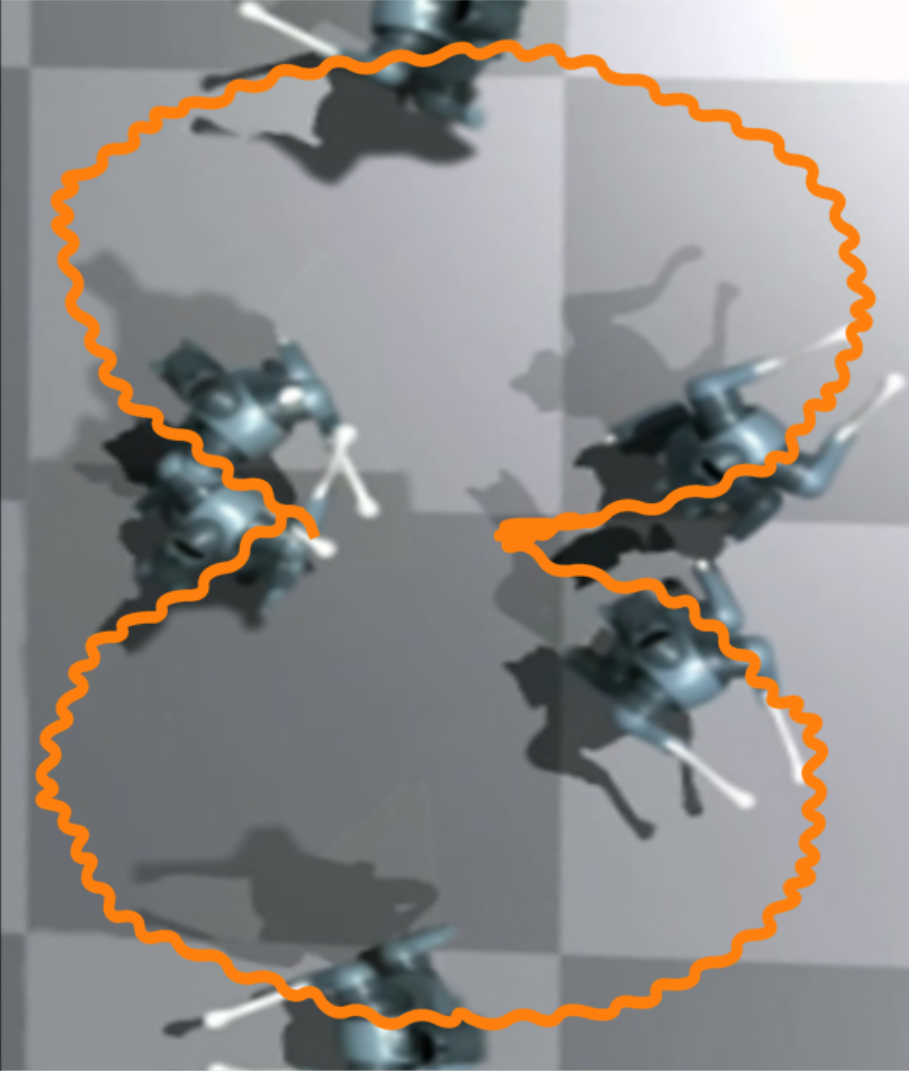}};
      \node[below=0pt of mlpplot] (mlp) {\includegraphics[width=0.435\linewidth, trim={110 0 0 400}, clip]{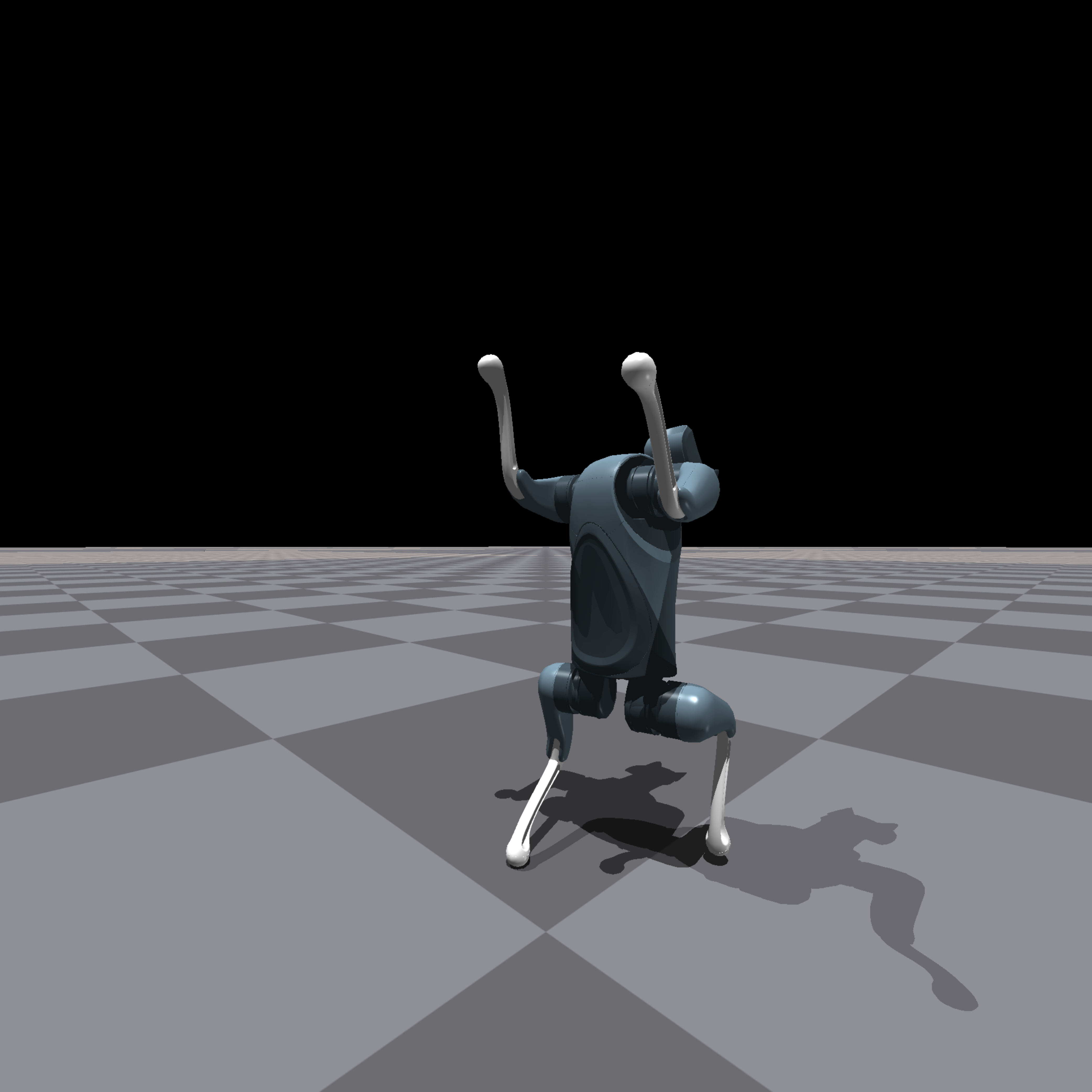}};
      \node[below=0pt of augplot] (aug) {\includegraphics[width=0.435\linewidth, trim={0 0 0 300}, clip]{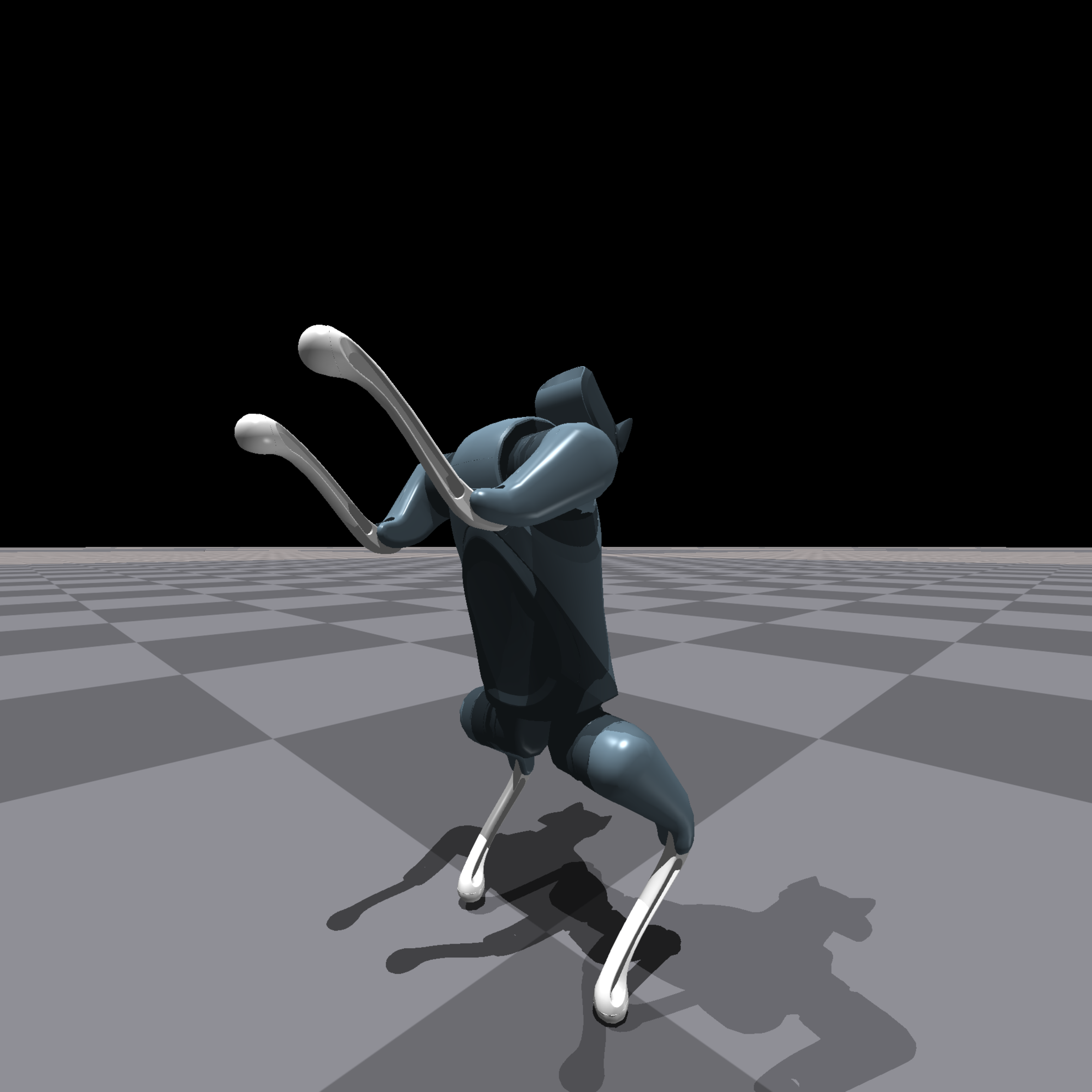}};
      \draw[my_blue] (mlpplot.north east) rectangle (mlp.south west);
      \draw[my_blue] (mlpplot.south east) rectangle (mlpplot.south west);
      \node[white, rectangle, fill=my_blue, inner sep=2pt, anchor=south west, align=center] at (mlp.south west) {\footnotesize Non-equivariant};
      \draw[my_green] (augplot.north east) rectangle (aug.south west);
      \draw[my_green] (augplot.south east) rectangle (augplot.south west);
      \node[white, rectangle, fill=my_green, inner sep=2pt, anchor=south west,align=center] at (aug.south west) {\footnotesize Equivariant};
    \end{tikzpicture}
    \caption{Comparison between non-equivariant (left) and equivariant (right) control policies, of a quadrupedal robot with a sagittal reflection symmetry, performing a right / left bipedal turning task. Top plots show the trajectories of a commanded turn in opposite directions. Bottom figures visualize the gait pattern the policy learns. While the unconstrained policy learns an asymmetric gait between left and right feet and fails to perform a symmetric turning trajectory, the equivariant policy achieves both motion-level and task-level symmetry. For more experimental results, see \url{https://youtu.be/Ad1clt4Yi4U}
    \vspace*{-.5cm}
    }
    \label{fig:stand-dance}
\end{figure}

Leveraging symmetries in \glspl{mdp} is a promising direction to alleviate the difficulty in symmetric-modality learning and provides a strong bias that we can leverage to improve the policy's exploration. Specifically, since symmetric \glspl{mdp} possess equivariant optimal control policies~\cite{wang2022robot,zinkevich2001symmetry_mdp_implications}, posting equivariance requirements on the RL policy helps it approach such an optimal policy. Since symmetry is ubiquitous in both biological and robotic systems, it is indeed an important but under-explored question of the best way to properly incorporate symmetry into RL algorithms. We note that although equivariance has gained attention in other fields such as computer vision and graphics, there have been limited prior works that apply it on legged robots and demonstrate its efficacy in related tasks.

In this work, we investigate different methods for adding equivariance information in improving the exploration guidance in training model-free RL algorithms. Specifically, we investigate how a loosely equivariant policy trained by data augmentation as carried out in~\cite{mittal2024symmetry} compare to a strictly equivariant policy enforced by its network architecture on legged robot control.  We perform extensive experiments in the challenging tasks of loco-manipulation and bipedal locomotion as quadrupeds. We benchmark the two methods against a vanilla RL policy in simulation environments and showcase the optimality of the symmetry-incorporated policies. Our experimental results show that the equivariance-enforced policy consistently outperforms other variants both in terms of the performance metrics and the acquired gait patterns being more steady and natural. 

We demonstrate the sim-to-real capability of both methods for completing a loco-manipulation task and a bipedal locomotion task, and show the enhanced robustness of the symmetry-incorporated policies compared to a vanilla RL policy. We provide a detailed discussion on the robustness of the two variants on real-world hardware, which could provide guidance for future development using symmetry-incorporated RL methods for different tasks.

\label{sec:intro}
% \cite{previousWork}

% ==================================================================
\section{Background}
% ==================================================================
\noindent
Here, we present a brief introduction to the fundamental concepts and notation required to understand how morphological symmetries can be leveraged in robot control.
% ------------------------------------------------------------------
\subsection{Symmetry group actions and representations}
% ------------------------------------------------------------------
\noindent
We will study the robot's morphological symmetry using the principles of group theory, a branch of mathematics that investigates symmetry transformations as abstract mathematical entities, separate from the objects they are linked with. This abstraction allows us to investigate how the robot's reflectional symmetry is imprinted on various vector spaces (such as the robot's state space and the \gls{mdp}'s action space) and functions of interest (including the robot's dynamics, and the \gls{mdp}'s value and reward functions).

Our focus is on the reflection symmetry group, denoted as $\G:=\CyclicGroup[2] = \{e, \g_s \;|\; \g_s^2 = e\}$. This group consists of the identity transformation, denoted by $e$, and the reflection transformation, denoted by $\g_s$, such that the transformation is its own inverse $\g_s \Gcomp \g_s := \g_s^2 = e$. 

To express the action of this group on a vector space $\vsX \subseteq \R^n$ we define a group representation on $\vsX$, denoted $\rep[\vsX]: \G \to \GLGroup(\vsX)$. This is a homomorphism (i.e., structure preserving) map between the symmetry transformations and the group of invertible linear maps on $\vsX$. The group structure is preserved in the sense that we can represent the composition of two symmetries by a matrix-matrix multiplication, $\rep[\vsX](\g_a \Gcomp \g_b) = \rep[\vsX](\g_a) \rep[\vsX](\g_b) \stforall \g_a,g_b \in \G$, and symmetry inversion by matrix inversion, $\rep[\vsX](\g^{\text{-}1}) = \rep[\vsX](\g)^{\text{-}1} \stforall \g \in \G$. This allows us to represent the action of a symmetry on a point $\vx \in \vsX$ as a matrix-vector multiplication, i.e., $\g \Glact \vx := \rep[\vsX](\g) \vx \in \vsX$. Here, $(\Glact): \G \times \vsX \to \vsX$ denotes the action of the group on the vector space. 

In this context, we will denote vector spaces that possess such a group action as \emph{symmetric vector spaces}. Next, we will define the symmetric properties of functions of symmetric vector spaces.

\subsection{Equivariant and invariant functions}
\noindent
A function $f:\vsX \rightarrow \vsY$ between two symmetric vector spaces typically falls into one of two categories: group invariant or group equivariant. If the output of the function $f$ remains constant regardless of the transformation applied to the input, the function is deemed $\G$-invariant. Conversely, a function is termed $\G$-equivariant if applying a transformation to the input before computing the function yields the same result as computing the function first and then applying the transformation to the output. These conditions can be formally defined as follows for all $\g \in \G, \vx \in \vsX$:
\vspace*{-.4cm}
\begin{multline}  
    \small
    \ubcolor[awesomeorange]{
        f(\vx) = f(\rep[\vsX](\g) \vx)
    }{\text{$\G$-invariant}}
    \;\; \text{and} \;\;
    \ubcolor[awesomeblue]{
        \rep[\vsY](\g) f(\vx) = f(\rep[\vsX](\g) \vx)
    }{\text{$\G$-equivariant}}
    .
    \label{eq:equivariance-invariance-constraints}
\end{multline}
\subsection{Symmetric Markov Decision Processes}
\label{sec:symmetric-mdp}
\noindent
An \gls{mdp} is defined by the tuple \((\vsS, \vsA, r, T, p_0)\), where \(\vsS\) is the state space, \(\vsA\) is the action space, \(r: \vsS \times \vsA \to \R\) is the reward function, \(T: \vsS \times \vsA \times \vsS \to [0, 1]\) is the transition density function, indicating the probability \(T(\vs' | \vs, \vaa)\) of transitioning to state \(\vs'\) given the current state \(\vs\) and action \(\vaa\). Lastly, \(p_0: \vsS \to [0, 1]\) is the probability density of initial states, denoting \(p_0(\vs)\) as the probability of starting the Markov process at state \(\vs\).

An \gls{mdp} is considered to have a symmetry group $\G$ if both state and action spaces are symmetric vector spaces (with the corresponding group representation $\rep[\vsS]$ and $\rep[\vsA]$), and if it satisfies the following conditions for all $\g \in \G, \vs, \vs' \in \vsS,\vaa \in \vsA$ \cite{wang2022robot,zinkevich2001symmetry_mdp_implications}:
\begin{subequations}
    \begin{itemize}
        \item The transition density is $\G$-invariant
    \end{itemize}
    % \vspace*{-.1cm}
    \begin{equation}
         T(\g \Glact \vs'| \g \Glact \vs, \g \Glact \vaa) = T(\vs'| \vs, \vaa)      
         \label{eq:g_inv_transition}
    \end{equation}
    % \vspace*{-0.7cm}
    \begin{itemize}
        \item The density of initial states is $\G$-invariant
    \end{itemize}
    % \vspace*{-.1cm}
    \begin{equation}
         p_0(\g \Glact \vs) = p_0(\vs)   
         \label{eq:g_inv_init_dist}
    \end{equation}
    % \vspace*{-0.7cm}
    \begin{itemize}
        \item The reward function is $\G$-invariant
    \end{itemize}
    % \vspace*{-.1cm}
    \begin{equation}
         r(\g \Glact \vs, \g \Glact \vaa) = r(\vs, \vaa)    
         \label{eq:g_inv_reward}
    \end{equation}
\end{subequations}

\noindent
\plaineqref{eq:g_inv_transition}, \plaineqref{eq:g_inv_init_dist} describe the necessary constraints on the \gls{mdp}'s dynamics to ensure that at every time $t$, the state probability density remains $\G$-invariant, $p_t(\g \Glact \vs) = p_t(\vs) \stforall t \geq 0, \g \in \G, \vs \in \vsS$. This characteristic is indicative of dynamical systems with $\G$-equivariant dynamics, implying that the temporal evolution of a given state $\vs$, on average, mirrors the evolution of its symmetric states $\g \Glact \vs  \stforall \g \in \G$ (refer to \cite{ordonezs2024morphological}).

The significance of studying such \glspl{mdp} lies in the known symmetry constraints of the optimal control policy $\pi^*: \vsS \to \vsA$ and the optimal value function $V^{\pi^*}: \vsS \to \R$. Specifically, symmetric \glspl{mdp} possess $\G$-equivariant optimal control policies \cite{zinkevich2001symmetry_mdp_implications},
% \vspace*{-.2cm}
\begin{equation}
    \g \Glact  \pi^*(\vs) = \pi^*(\g \Glact \vs) \stforall \vs \in \vsS, \g \in \G,
\label{eqn:g_policy}
\end{equation}
% \vspace*{-.5cm}
and $\G$-invariant optimal value functions \cite{zinkevich2001symmetry_mdp_implications},
% \vspace*{-.2cm}
\begin{equation}
    V^{\pi^*}(\g \Glact \vs) = V^{\pi^*}(\vs)  \stforall \vs \in \vsS, \g \in \G.
    \label{eqn:g_value}
\end{equation}
% \vspace*{-.85cm}
\noindent
These constraints provide substantial inductive biases that can be leveraged in reinforcement learning algorithms  \cite{wang2022robot,huang2022equivariant,van2020mdp}. As we will discuss, the sagittal symmetry inherent in a quadruped robot's morphology allows us to formulate the controlled robot dynamics as a symmetric \gls{mdp}.

\subsection{Morphological symmetries and symmetric \glspl{mdp}}
\label{sec:ms_symm_mdps}
\noindent
In this work we aim to learn loco-manipulaiton control policies for the quadruped robot Xiaomi CyberDog2 \cite{cyberdog2} and Unitree Go1 \cite{unitree2024website}. These robots possess a sagittal state symmetry arising from the symmetric mass distribution of the robot's torso, and the replication of the left-right limbs. 

This symmetry, referred to as a morphological symmetry, enable us to cast the robot's controlled dynamics as a symmetric \gls{mdp}, considering that, due to the morphological symmetry, the robot's state vector space, denoted $\vsS \in \R^{n}$, and action space, denoted $\vsA \in \R^{n_a}$, are symmetric vector spaces \cite{ordonezs2024morphological,ordonez2023discretesymm}, with the action space being spanned by the robot's $n_a \leq n$ controlled \gls{dof}. Furthermore, as robots with morphological symmetries feature $\G$-equivariant dynamics \cite{ordonezs2024morphological}, the \gls{mdp}'s transition density $T$ is guaranteed to be $\G$-invariant, complying with \plaineqref{eq:g_inv_transition}. Moreover, the required $\G$-invariance of the reward function $r$ \plaineqref{eq:g_inv_reward} is commonly satisfied as the reward depends only on relative distance/error measurements of the robot state, and other $\G$-invariant terms, such as fall/contact detection. 

\section{Related Works}
\label{sec:related-works}
\subsubsection*{Leveraging Symmetry in RL}
\noindent 
Previous works on leveraging symmetry in locomotion control learning focus on leveraging morphological symmetries \cite{ordonezs2024morphological} and / or temporal symmetries \cite{ding2023breaking} (i.e., periodicity of locomotion gait). 

To encourage periodic gaits, \cite{liu2016guided,abdolhosseini2019symmetric_locomotion} introduce periodic phase signals in the state space $\vsS$, characterizing the gait cycle and / or each of the limb's gait cycle. Alternatively, \cite{lee2020learning} defines the \gls{mdp}'s action space $\vsA$ in terms of the parameters of existing Central Pattern Generators \cite{bellegarda2022cpg} describing periodic motion of the robot's limbs. 

While these methods only explore temporal symmetry, in this work we present a method incorporating both temporal and morphological symmetries. 
Leveraging morphological symmetries to aid in the approximation of the optimal $\G$-equivariant control policy $\pi^{*}$ \plaineqref{eqn:g_policy} and $\G$-invariant value function $V^{\pi^{*}}$ \plaineqref{eqn:g_value} can be achieved either through data-augmentation and soft / hard equivariance / invariance constraints on the models used to approximate these functions.

Inspired by data augmentation commonly used in supervised learning, augmenting RL algorithms with symmetric data involves constructing gradient from both the collected transitions and their symmetric transitions to induce equivariance / invariance for the policy and value function.  This simple idea proves to be effective in locomotion tasks \cite{mittal2024symmetry} and manipulation tasks~\cite{lin2020invariant}. 

% Loss 
Modifying loss functions to regulate equivariance in the policy and invariance in the value function provides additional gradient with symmetry information and poses \textit{soft} symmetry constraints for the RL policy. Prior works show that an effective auxiliary objective for the policy network can be simply adding a regularization loss for equivariant actions~\cite{yu2018learning}. However, this requires delicate hyperparameter tuning. 
Work in \cite{kasaei2021cpg} adds a trust region on the regularization loss to improve training stability, and \cite{ordonez2022adaptable}
adapts it to Soft Actor Critic algorithm and adds regularization on the variance of the action distribution.

% Equivariant Neural Networks
Enforcing $\G$-equivariant policies and $\G$-invariant value functions involves modifying neural network model architecture~\cite{mondal2020group, rezaei2022continuousMDPhomomorphisms, van2020mdp}, which introduces \textit{hard} constraints on the symmetry of the learnt policy. In manipulation tasks, prior works~\cite{wang2022mathrm, wang2022robot} have shown that such hard equivariance constraints lead to superior performance among other soft, approximately equivariant policies. However, the applicability of such equivariant RL algorithm and its performance on more dynamic robotics systems, such as legged robots remain underexplored. In this work, we study the effectiveness of data augmentation and hard-constraint RL algorithms on challenging locomotion and loco-manipulation tasks. 

\subsubsection*{Loco-manipulation and  Bipedal Locomotion}
\noindent
Recent works have focused on pushing the boundary of agility of legged robots beyond locomotion tasks. One trend is the demonstration of loco-manipulation tasks with legs being the end-effector to interact with objects, such as buttons on the wall \cite{cheng2023legs, vollenweider2023advanced}, soccer balls \cite{ji2023dribble, huang2023creating}, and move environment objects as simple tools \cite{xu2023creative}. However, as the task becomes more complex and requires highly dynamic maneuvers, it becomes exponentially more difficult for the policy to learn symmetrical behavior, resulting in suboptimal performance as seen in \cite{huang2023creating} and \cite{ji2023dribble}. 

Another agile task commonly demonstrated is to perform bipedal locomotion with quadrupedal robots. Prior works demonstrate bipedal walking with assistive devices to lower the robustness requirements~\cite{yu2022multi}, or show single-task bipedal walking without capabilities to effectively turn and walk at various velocities~\cite{smith2023learning}. A recent work shows a more developed controller, but again it shows non-negligible asymmetrical performance in agile motions, especially turning in various directions~\cite{li2023learning}. For humanoid robots, we again observe asymmetric behaviors on agile skills such as jumping to the side~\cite{li2023robust} when trained without equivariance constraints. In this work, we aim to investigate how adding symmetry constraints to the RL algorithms may affect the performance on such complex and agile bipedal locomotion tasks.

%% RL
% 
% 
\section{Incorporating Symmetry in Model-free RL} \label{sec:RL}
\noindent
In this section, we introduce two variations of \gls{PPO}~\cite{schulman2017proximal} which distinctly leverage the robot's morphological symmetry, namely: \gls{PPO} with data-augmentation in training (\textbf{\acrshort{PPOaug}}) and 
\gls{PPO} with hard equivariance/invariance constraints on neural networks (\textbf{\acrshort{PPOsymm}}). We will compare these variants with a vanilla \textbf{\gls{PPO}} implementation as a baseline. We opt out symmetric loss function, another method to augment the training process, as it has been consistently outperformed by \acrshort{PPOaug}, as prior work~\cite{mittal2024symmetry} shows. 
To build the symmetry representations $\rep[\vsS]$ and $\rep[\vsA]$, we leverage open-sourced repositories MorphoSymm~\cite{ordonezs2024morphological} and ESCNN~\cite{cesa2021program}. 

\subsection{\acrfull{PPOaug}}

\noindent
Considering the symmetries of the state and action spaces, and the invariance of the reward \plaineqref{eq:g_inv_reward}, a direct method to leverage symmetry is data-augmentation on policy and critic learning. Specifically, we perform updates on policy $\pi_\theta$ and critic $V_\phi$, parameterized by $\theta$ and $\phi$, with both the augmented transition tuples $(\g \Glact \vs', \g \Glact \vaa, \g \Glact \vs, r(\g \Glact \vs, \g \Glact \vaa))$ and the online collected transition tuples $(\vs', \vaa, \vs, r(\vs, \vaa))$. 

Data augmentation in learning the critic biases the learned value function to be an approximately $\G$-invariant function. This, in turn, guides the optimization of the actor control policy, by constructing an invariant return estimate within the PPO's general advantage estimation. Following 
\plaineqref{eq:g_inv_reward}, the augmentation process ensures that the values of the symmetric states converge towards identical target returns.
% thereby biasing the function to become approximately $\G$-invariant.

Since \gls{PPO} is an on-policy algorithm, the fact that the augmented sample does not strictly come from the action distribution of the current policy creates an off-policy scenario. 
% to an on-policy algorithm. Concretely, this means that for policy distribution $\pi$, $\pi(\g \Glact \vaa | \g \Glact \vs) \neq \pi(\vaa | \vs)$ for original and augmented state-action pairs. improve stability in \gls{PPO} training and
To minimize the effects of off-policy samples, we need to ensure that policy $\pi_\theta$ is approximately symmetric.
% , that is, $\pi(\g \Glact \vaa | \g \Glact \vs) \approx \pi(\vaa | \vs)$. 
To achieve this, we seek to minimize the difference between the gradient signals between the original samples and the augmented samples by network initialization and update rules. Specifically, we initialize the policy network with zero mean $\mu = 0$ and a small variance such that for the initialized policy network, $\pi_\theta(\g \Glact \vs) \approx \pi_\theta(\vs)$ with a large enough standard deviation for samples in the first update. Then, unlike previous work \cite{lin2020invariant} which stores the equivariant transitions in the rollout storage, we opt to perform equivariant augmentation after sampling a mini-batch of original data for each gradient update, such that the loss considers the original and augmented samples equally. 
% Specifically, for gradient step $n$, we expect $\g \Glact (\nabla \pi(\g \Glact \vaa_n | \g \Glact \vs_n)) \approx \nabla \pi(\vaa_n | \vs_n)$ given $\pi(\g \Glact \vaa_n | \g \Glact \vs_n) \approx \pi(\vaa_n | \vs_n)$. 
Assuming that the policy $\pi_\theta$ before update is approximately equivariant, since the augmented samples provide gradient signals that push the policy towards an equivariant action distribution to a similar extent alone with the original one, we expect the updated policy to be approximately equivariant as well. 
% since \eqref{eq:g_reward} holds for symmetric MDP, the policy gradient $PG = r(s,a)\nabla log_{\pi(\vaa|\vs)}$ of the equivariant samples is approximately symmetric to that of the original samples. 
% For the value function, since \eqref{eq:g_inv_reward} holds, the values of the equivariant state regress to the same target returns, such that the value function is also approximately $\G$-invariant. 
% Empirically we find this formulation leads to an approximately equivariant policy that can be trained stably with PPO, as shown in Sec. \ref{sec:training}.  

\subsection{\acrfull{PPOsymm}}
\noindent
To enforce $\G$-equivariance constraints on the learned control policy $\pi_\theta$ and $\G$-invariance constraints on the learned value function $V_\phi$, we parameterize these functions as equivariant / invariant neural networks. In this work, we design these networks as equivariant / invariant multi-layer-perceptron (EMLP) as a substitute for the unconstraint MLP. 
Specifically, EMLP network guarantees that $
 \pi_\theta(\g \Glact \vs) = \g \Glact  \pi_\theta(\vs)$. 
% $\pi(\g \Glact \vaa | \g \Glact \vs) = \pi(\vaa | \vs) \stforall \vaa, \vs \in \vsS, \vsA$. 
Regarding the invariant value function, we substitute the critic MLP with an invariant MLP, such that for an invariant NN, $V_\phi(\g \Glact \vs) = V_\phi(\vs)$. This is done by swapping the last layer of the EMLP with an invariant transformation layer that transforms the regular representation of symmetry group $\G$ into the trivial representation, which is a scalar for the reflection group $\CyclicGroup[2]$. We train the equivariant actor and invariant critic networks with vanilla \gls{PPO} algorithm. 

Besides morphological symmetry, we further consider temporal symmetry by assuming gait periodicity, leading to a valid equivariance transformation on the phase signal $\psi$~\cite{ding2023breaking}, as in Sec. \ref{sec:related-works}. Specifically, for the phase signal $\psi$ that varies from 0 to 1 within each period, the operation $\g \Glact \psi$ yields $(\psi + 0.5) - \lfloor \psi + 0.5 \rfloor$.
A potential problem previous works have noted is that equivariant policies might struggle with neutral states~\cite{abdolhosseini2019symmetric_locomotion, mittal2024symmetry} where $\g \Glact \vs = \vs$ leading to $\g \Glact \vaa = \vaa$, for all $\g \in \G$. By incorporating the additional phase $\psi$ into the state of the MDP, we let $\g \Glact \vs \neq \vs$ when the robot is in its morphological neutral phase, thus avoiding the problem of neutral states for equivariant policy.

%% TASKS DESCRIPTION
\section{Experiments}
\noindent
We compare the \gls{PPO} variants described in Sec.~\ref{sec:RL} on four different tasks consisting of loco-manipulation and bipedal locomotion on quadrupedal robots. 
In loco-manipulation tasks, we investigate how incorporating symmetry influences task-level symmetry, while in bipedal locomotion, we focus on the intrinsic symmetry from the quadruped kinematic and dynamics structure.
The observation consists of proprioceptive readings and task-specific observations. For tasks involving bipedal walking, we give dense rewards that encourage up-right orientation of the base and feet stepping periodically, adapted from \cite{li2023learning}. 
All rewards are $\G$-invariant functions with relative measurements of the robot state. 
We include domain randomization for easing the sim-to-real transfer. All the tasks are trained in the Isaac Gym simulator~\cite{makoviychuk2021isaac}. Details of the environment setup are described below. 

\begin{figure}[t]
    \centering
    \def\smallgap{1pt}
    \begin{tikzpicture}[inner sep=0pt]
        \node (pushdoor) {\includegraphics[width=0.24\linewidth]{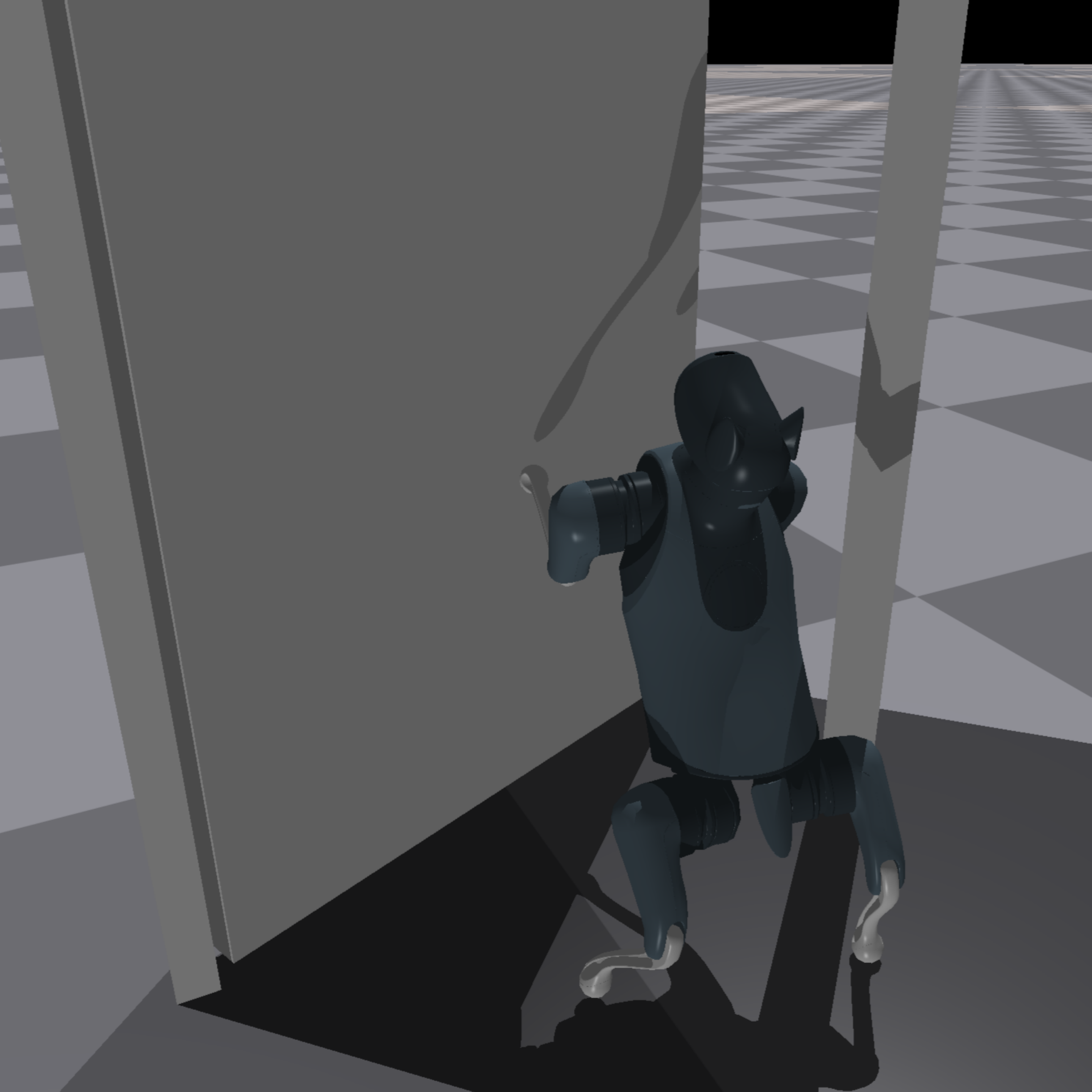}};
        \node[right=\smallgap of pushdoor] (dribbling) {\includegraphics[width=0.24\linewidth]{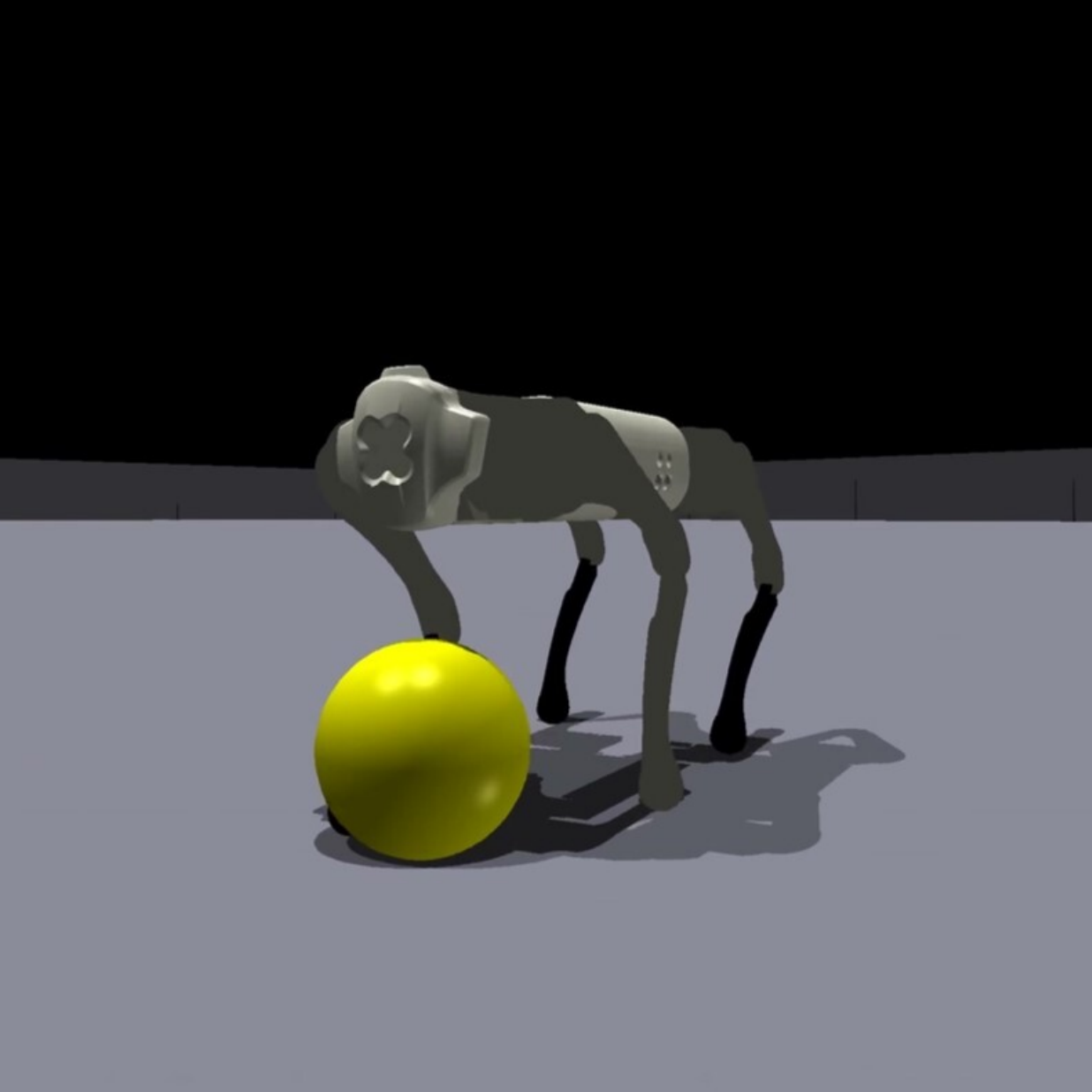}};
        \node[right=\smallgap of dribbling] (standdance) {\includegraphics[width=0.24\linewidth]{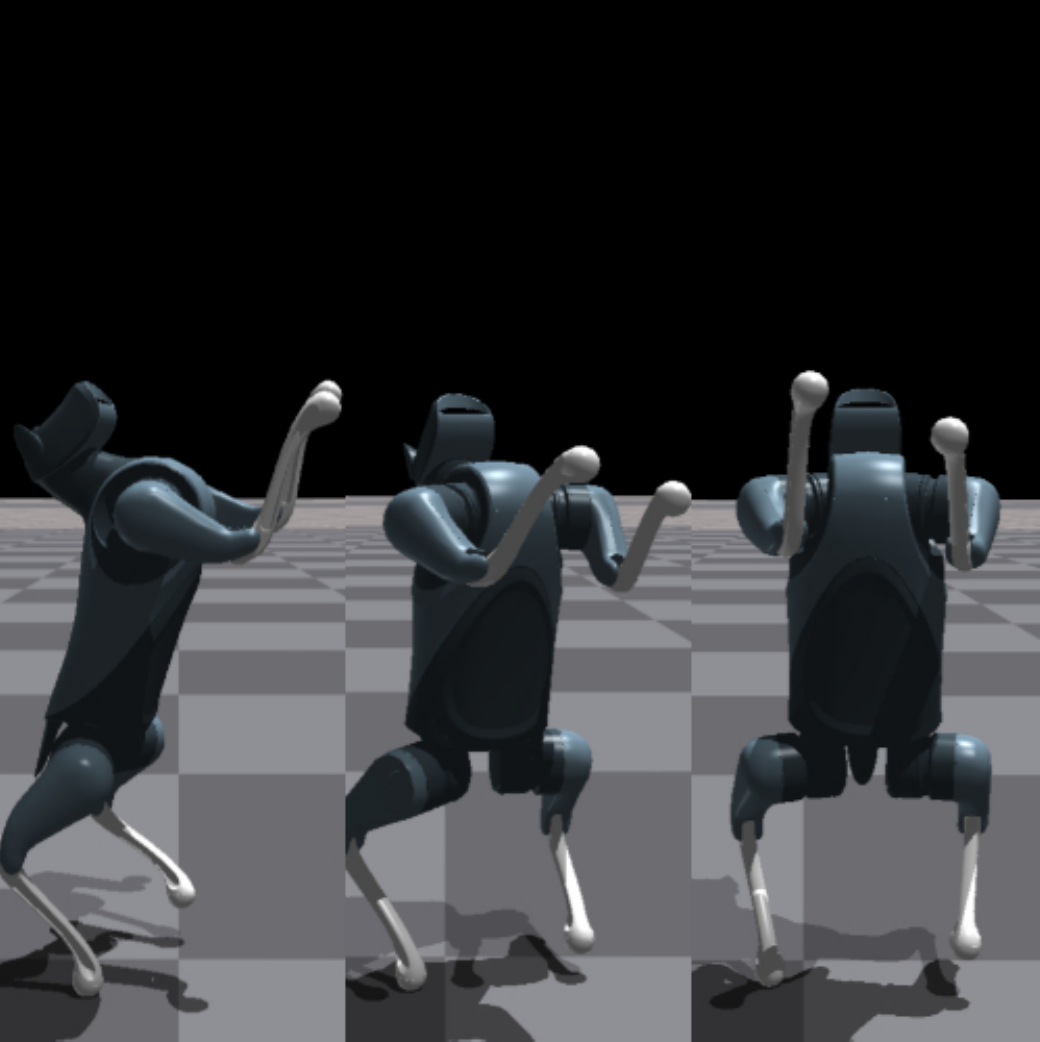}};
        \node[right=\smallgap of standdance] (slopwalking) {\includegraphics[width=0.24\linewidth]{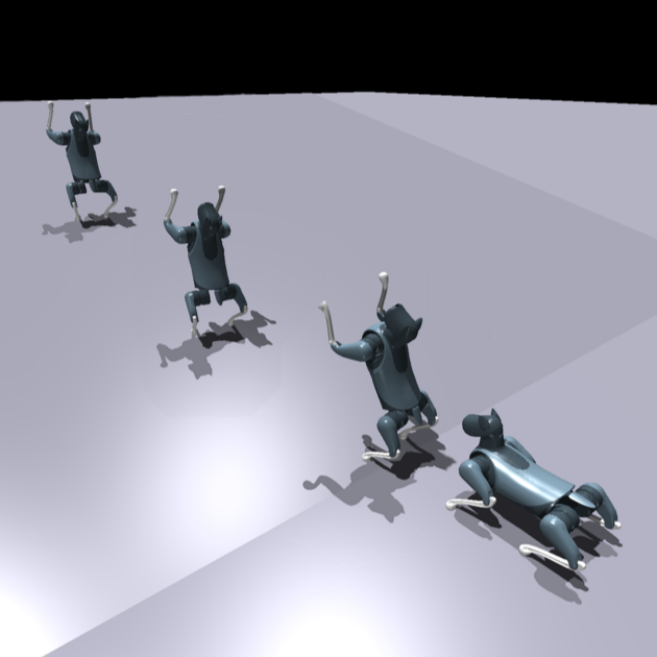}};
        \node[white, anchor=north west, inner sep=1pt] at (pushdoor.north west) {\scriptsize (a) Door Pushing};
        \node[white, anchor=north west, inner sep=1pt] at (dribbling.north west) {\scriptsize (b) Dribbling};
        \node[white, anchor=north west, inner sep=1pt] at (standdance.north west) {\scriptsize (c) Stand Turning};
        \node[white, anchor=north west, inner sep=1pt] at (slopwalking.north west) {\scriptsize (d) Slope Walking};
    \end{tikzpicture}
\caption{Challenging tasks to test the efficacy of symmetry-incorporated policies. (a,b) showcase loco-manipulation with task-space symmetry and (c,d) exhibit bipedal locomotion with motion-level symmetry. These tasks push the boundary of agility and dynamic mobile manipulation for quadrupeds.
\vspace*{-.5cm}
}
\label{fig:tasks}
\end{figure}
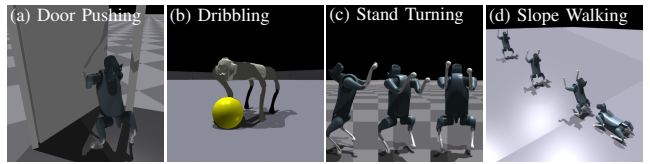

\textbf{Door Pushing:} \label{task:push door}
% This loco-manipulation task (\ref{fig:tasks}(a)) involves a robot standing up and opening a door with its front limbs, adjusting its orientation for doors that open to the right or left. The task-specific observations include the initial state of the door (relative position with respect to the door, relative normal vector, opening direction) in the robot's frame. The task specific reward includes a forward velocity tracking error only, as the policy is expected to learn to push the door to move forward. 
In this loco-manipulation task (Fig.~\ref{fig:tasks}(a)), a robot needs to open a door using its front limbs while standing and adjust for doors that open either left or right. The task-specific observations include its initial relative position and orientation with respect to the door in the robot's frame, and the door's swing direction. The task-specific reward only includes the tracking error in forward velocity, encouraging the policy to learn pushing the door open to advance.

\textbf{Dribbling:} \label{task:dribbling}
Shown in Fig. \ref{fig:tasks}(b), this task requires a quadrupedal robot to dribble a soccer ball given its desired velocity commands. The ball observation is a relative position with respect to the robot base frame, and the rewards include ball velocity tracking and robot's proximity to the ball. We refer readers to \cite{ji2023dribble} for more detailed settings. 

\textbf{Stand Turning:} \label{task:stand turn}
This agile bipedal locomotion task, adapted from~\cite{li2023learning}, requires a quadrupedal robot to stand up on two feet and follow input commands, consisting of the desired linear velocity and the robot's desired heading, while the tracking errors are used as task rewards. 

\textbf{Slope Walking:} \label{task:walk slope}
This bipedal locomotion task pushes the limit of agility by requiring a quadruped to traverse an inclined flat surface on two feet.Training involves a terrain curriculum with a maximum slope of $11.3^\circ$. 

%% RESULTS
\section{Results}
\noindent
In this section, we discuss the training and evaluation of PPO, PPOaug, and PPOeqic on the aforementioned tasks. We keep the training scheme the same across three methods, and we tune the hyperparameters separately for each task. For quantitative results, we evaluate on 2,000 simulation environments and 10 episodes per environment. We report the mean and variance across three different seeds. 

\subsection{Training Performance}
\label{sec:training}
\noindent
For training performance, we focus on both training return (highest return in training) and sample efficiency (number of samples needed to reach the same return) as the metrics. 

\begin{figure*}[t]
  \centering
  \def\width{0.24\linewidth}
  \def\smallgap{0pt}
  \begin{tikzpicture}[inner sep=0pt]
      \node (doorpushing) {\includegraphics[width=\width]{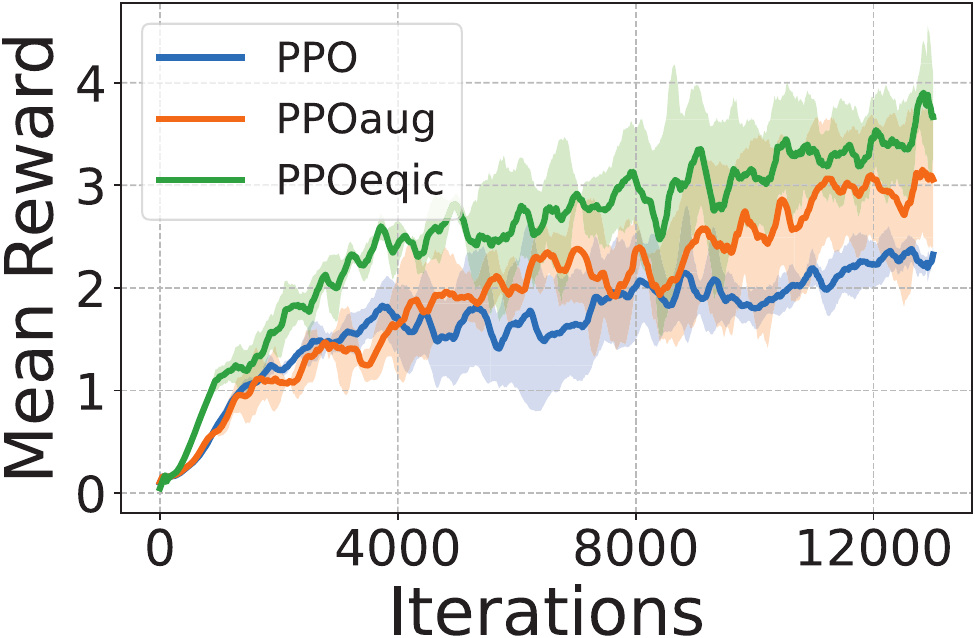}};
      \node[right=3pt of doorpushing,] (dribbling) {\includegraphics[width=\width]{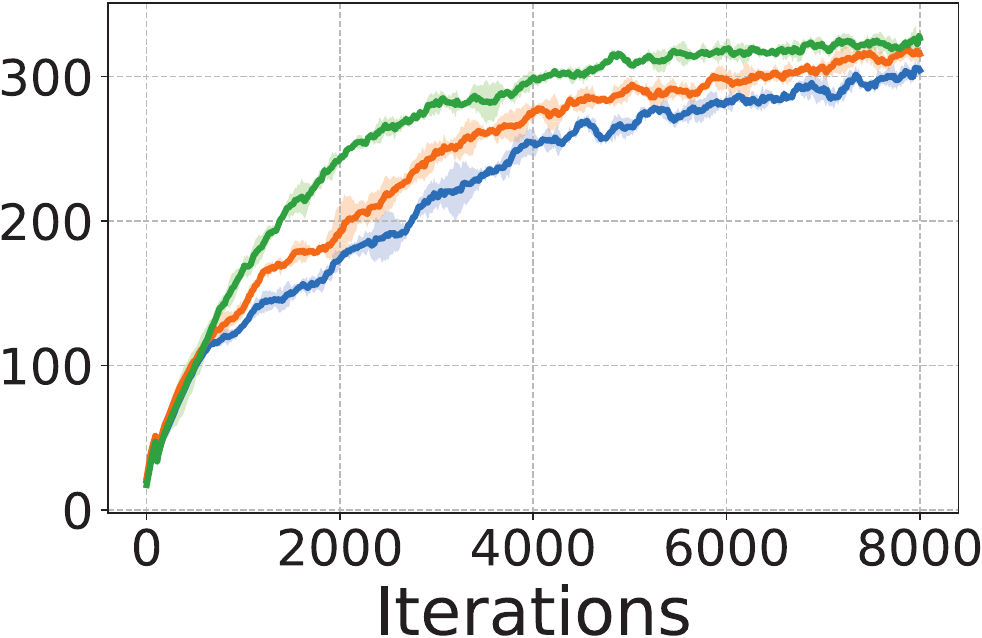}};
      \node[right=2pt of dribbling] (standdance) {\includegraphics[width=\width]{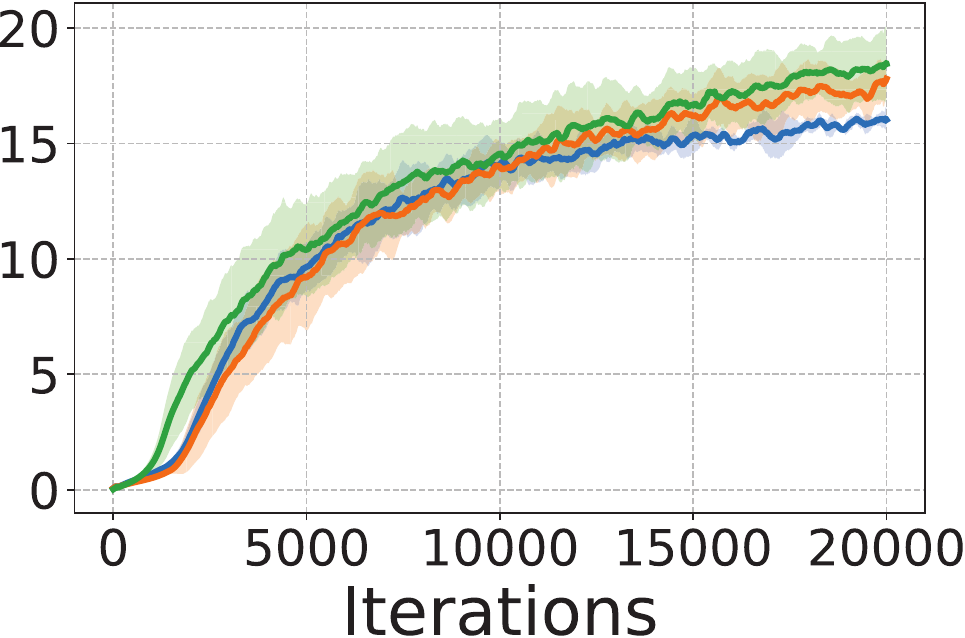}};
      \node[right=\smallgap of standdance] (walkslop) {\includegraphics[width=\width]{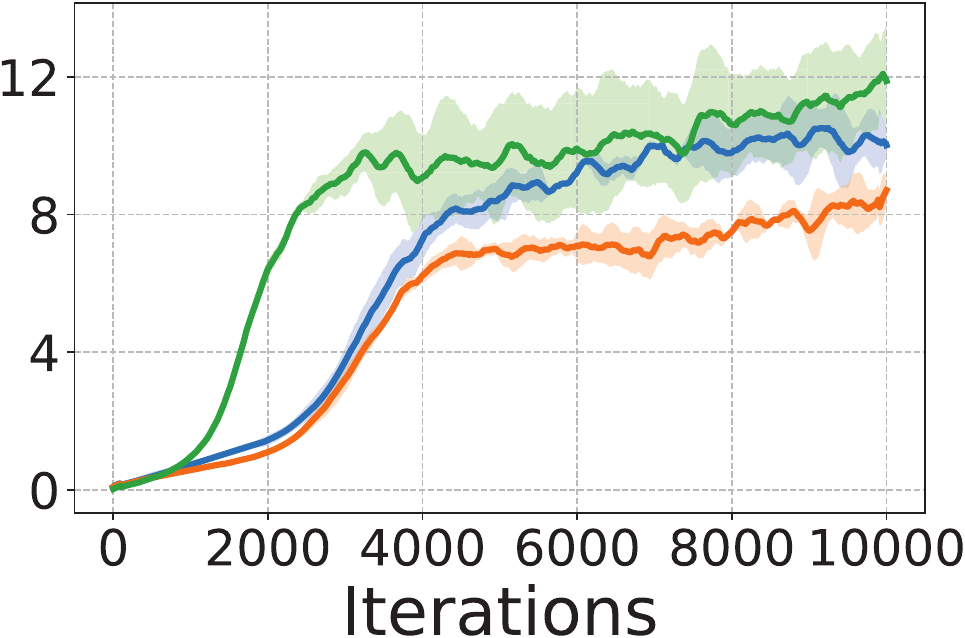}};
  \end{tikzpicture}

\caption{Comparison of training curves of PPO, \gls{PPOaug}, and \gls{PPOsymm} on four tasks from left to right: Door Pushing, Dribbling, Stand Turning, and Slope Walking. Learning curves show mean episodic return and standard deviation for three seeds. \gls{PPOsymm} consistently demonstrates the highest training returns and sample efficiency in all tasks.}
\label{fig:exp-bcloss}
\end{figure*}

From Fig. \ref{fig:exp-bcloss}, the \gls{PPOsymm} policy consistently outperforms other \gls{PPO} variants, showing the effectiveness of equivariance constraints in training. 
Furthermore, \gls{PPOsymm} shows consistent improvements in sample efficiency, notably with fewer samples in early stage of training. 

This indicates the equivariance constraints provide more efficient guidance in the policy's exploration than other variants. 
In contrast, \gls{PPOaug} achieves higher training returns but similar sample efficiency across three out of four tasks compared to vanilla \gls{PPO} baseline. 
Unlike \gls{PPOsymm}, \gls{PPOaug} needs to learn two aproximately equivariant action distributions, i.e. both $\pi_\theta(\g \Glact \vs)$ \textit{and} $\pi_\theta(\vs)$, which is itself a more complex learning scheme. In comparison, vanilla \gls{PPO} tends to converge to an inequivariant policy that still receives high training returns, such as leveraging an asymmetric gait. 

Notice that these sub-optimal behaviors appeared on the \gls{PPO} policy, though feasible due to imprecise contact modeling, are overfitted to the simulation environment and lack robustness, which are proved to be infeasible in sim-to-real transfer, as later described in Sec. \ref{sec:real-world}. 

\subsection{Door Pushing task}
\label{sec:push-door}
\begin{table*}
\centering
\begin{tabular}{llcccccc}
\toprule
\textbf{Method} & \textbf{} & \textbf{Mean SR ($\%$)} & \textbf{Max SR ($\%$)} & \textbf{RSI} & \textbf{OOD Mean SR ($\%$)} & \textbf{OOD Max SR ($\%$)} & \textbf{OOD RSI}\\
\midrule
\textbf{PPO}
    & trained on 1 side & 43.40 $\pm$ 1.73 & 44.47 & 199.96 & 27.46 $\pm$ 2.34 & 30.04 & 199.99 \\
    & trained on 2 sides & 61.18 $\pm$ 7.56 & 69.63 & 12.25 & 42.98 $\pm$ 2.21 & 45.51 & 3.45 \\
    % & $\|\omega\|$ (\unit{rad / s}) & 0.12 $\pm$ 0.02 & 0.26 $\pm$ 0.01 & 0.20 $\pm$ 0.05 & 0.46 $\pm$ 0.01 & 0.45 $\pm$ 0.03 \\

% \addlinespace
\midrule
% \addlinespace

\textbf{\gls{PPOaug}}
    & trained on 1 side & 54.39 $\pm$ 32.56 & 86.98 & 3.02 & 38.24 $\pm$ 19.17 & 52.85 & 1.40\\
    & trained on 2 sides & 50.24 $\pm$ 36.52 & 74.98 & 3.77 & 36.46 $\pm$ 25.98 & 53.78 & \textbf{0.17}\\
    % & $\|\omega\|$ (\unit{rad / s}) & 0.48 $\pm$ 0.01 & 0.42 $\pm$ 0.01 & 0.48 $\pm$ 0.01 & 0.74 $\pm$ 0.04 & 0.62 $\pm$ 0.01 \\

% \addlinespace
\midrule
% \addlinespace

\textbf{\gls{PPOsymm}}
    & trained on 1 side & \textbf{65.65} $\pm$ 23.16 & \textbf{87.96} & \textbf{0.98} & \textbf{44.15} $\pm$ 9.39 & 
 50.63 & 0.88\\
    & trained on 2 sides & 59.92 $\pm$ 17.39 & 74.74 & 1.65 & 38.56 $\pm$ 15.54 & \textbf{55.95} & 0.32\\
    % & $\|\omega\|$ (\unit{rad / s}) & 0.81 $\pm$ 0.10 & 0.82 $\pm$ 0.14 & 0.88 $\pm$ 0.08 & 1.06 $\pm$ 0.18 & 0.82 $\pm$ 0.01 \\

% \addlinespace
\bottomrule
% \addlinespace

\end{tabular}
\caption{\label{tab:pushdoor-performance-metrics} Comparison of success rates (SR) and their symmetry index on door-pushing tasks on training-distribution and out-of-distribution scenarios. Of the three variants, \gls{PPOsymm} demonstrates both higher success rate and better symmetry index in both cases, indicating a better task-level symmetric policy. } 

\end{table*}

\noindent
We select the door pushing task as the benchmark for task-level symmetry in loco-manipulation, we evaluate the performance using two key metrics: success rate (SR) and the Symmetric Index (SI)~\cite{robinson1987use}. SR quantifies the percentage of episodes where the robot is able to open a door at least 60 degrees wide and traverse it successfully, whereas SI assesses the disparity in success rates between left and right scenarios, calculated as $\text{SI} = \frac{2|X_R-X_L|}{X_R+X_L}\times 100\%$, where $X_R$ is a scalar metric (e.g. the success rate) and $X_L$ is its symmetric version. 

\subsubsection{Success Rates}
As shown in Table \ref{tab:pushdoor-performance-metrics}, \gls{PPOsymm} policy consistently outperforms the unconstrained \gls{PPO} policy with a $4.47\%$ higher mean SR. \gls{PPOaug}, in comparison, achieves a $6\%$ lower mean SR. However, in addition to mean success rates, we are also interested in the \textit{maximum} success rate each method may achieve across the three seeds. We notice that the difference between mean and maximum SRs for \gls{PPO} policy is small, indicating that it consistently converges to similar sub-optima. In comparison, both symmetry-incorporated policies achieve a much higher maximum SR, around $88\%$ for \gls{PPOsymm} and $87\%$ for \gls{PPOaug}, which is $19\%$ higher than that of \gls{PPO} policy. This indicates that while there is no guarantee, \gls{PPOsymm} and \gls{PPOaug} are indeed more likely to achieve a better performance than PPO, given the optimal policy for the task is equivariant. 
% Thus, even though the mean SR for \gls{PPOaug} policy is lower, we do not conclude that \gls{PPOaug} does not benefit from symmetry constraints. 

Comparing \gls{PPOaug} and \gls{PPOsymm}, we notice that while they achieve similar maximum SRs, the mean SR for \gls{PPOsymm} is $10\%$ higher than the one of \gls{PPOaug}. This showcases the robustness in training of the hard-constraint equivariant policy compared to the soft-constraint policy.

\subsubsection{Symmetry Index}
Looking at the SI metric, we find that unconstrained \gls{PPO} policy receives a high SI value, 4 times higher than the \gls{PPOaug} policy and 8 times higher than the hard-constraint \gls{PPOsymm} policy. 
This indicates that although \gls{PPO} achieves a relatively high SR, it fails to learn a balanced policy between the two symmetric modes (left and right-handed doors). 
% As the RL policy is commonly modeled as a single-mode distribution, as is in our case, it is hard for the policy to learn both modes well. 
As previously mentioned, vanilla RL faces exploration difficulty in learning multi-modal distributions, such as the two modes in our case.
In comparison, both \gls{PPOaug} and \gls{PPOsymm} policy achieve low SIs, indicating symmetric behaviors in the \textit{task} space, even though the equivariance constraints are on the \textit{joint} space.

\subsubsection{Out-of-Distribution Scenarios}
The policy's performance is further evaluated on its robustness and generalization to \gls{ood} scenarios. Specifically, we randomize the robot's initial roll, pitch, and yaw angles in a wide range of [$-15^{\circ}, 15^{\circ}$] that is unseen during training. 

In out-of-distribution scenarios, similar to in-distribution results, \gls{PPOsymm} outperforms the vanilla PPO policy in both mean and maximum SRs. 
In addition, \gls{PPOsymm} again demonstrates more stable training performance than \gls{PPOaug} policy, achieving a $5.9\%$ higher mean SR across three seeds. This showcases the better generalization of an equivariant policy in response to out-of-distribution scenarios, given the distribution still follows a symmetric MDP setting.

\subsubsection{Single Mode Training}
An intriguing finding is that while \gls{PPOsymm} has hard equivariance constraint, the policy trained only with one side scenario consistently outperforms the one trained on both sides. 
This trend is also present in \gls{PPOaug}, even though the policy is not equivariance-enforced. 
We attribute this to the fact that during training, the domain randomization of joint properties, such as friction and PD gains, breaks the assumption of $\G$-invariant state transition probability that is necessary for an optimal policy to be equivariant. 
Consider a case where the joint properties are asymmetrically randomized, if the same environment is trained on only right-handed door, since the direction indicator is a part of the state $\vs$, for any state transition $(\vs,\vaa,\vs')$, we can mirror an equivariant transition $(\g \Glact \vs, \g \Glact \vaa, \g \Glact \vs')$ that satisfies \eqref{eq:g_inv_transition} and does not conflict with the training data. 
Essentially, we can mirror a symmetric environment not seen in training but exists in the training distribution to form a symmetric MDP together with the original environment. 
However, if $(\g \Glact \vs, \g \Glact \vaa, \g \Glact \vs')$, the mirrored transition with the left-handed door already exist in the training data, 
\eqref{eq:g_inv_transition} will not hold because of the asymmetric dynamics. 
Therefore, in contexts requiring dynamic asymmetry, such as domain randomization, it is beneficial to train on only one mode but deploy on other symmetric modes with equivariant policies. As we show in Sec. \ref{sec:real-world}, \gls{PPOsymm} trained with only right-handed doors can be zero-shot deployed on doors that swing on either side.

\subsection{Dribbling task}
\noindent
% Consistent with previous loco-manipulation task, we observe that incorporating symmetry helps improve soccer dribbling gait. Qualitatively, we observe that \gls{PPOsymm} and \gls{PPOaug} tend to keep the ball closer to its body, while \gls{PPO} often kicks ball away and chases it. Quantitatively, the hard-constraint \gls{PPOsymm} policy gets an average episodic return of $431.86 \pm 1.77$, narrowly outperforming \gls{PPO} ($427.02 \pm 2.15$) by 1.1\% and \gls{PPOaug} ($430.49 \pm 3.45$), indicating a minor lead possibly because all methods saturate in the simulation environment.

Consistent with previous loco-manipulation tasks, incorporating symmetry improves soccer dribbling gaits. Qualitatively, \gls{PPOsymm} and \gls{PPOaug} keep the ball closer to the robot, while \gls{PPO} often kicks it away and chases it. Quantitatively, the hard-constraint \gls{PPOsymm} achieves an average episodic return of $431.86 \pm 1.77$, slightly outperforming \gls{PPO} ($427.02 \pm 2.15$) and \gls{PPOaug} ($430.49 \pm 3.45$), suggesting all methods are near saturation in the simulation environment.

\subsection{Stand Turning task}
\begin{table}

\centering
\begin{tabular}{lcccc}
\toprule
\small
\textbf{Method} & \textbf{Error (rad)} & \textbf{Error RSI} & \textbf{CoT (Js/m)} & \textbf{CoT RSI} \\
\midrule
\textbf{PPO}
    & 0.265 $\pm$ 0.022 & 0.0945 & 2223 $\pm$ 102 & 0.0259  \\
    % & $\|\omega\|$ (\unit{rad / s}) & 0.12 $\pm$ 0.02 & 0.26 $\pm$ 0.01 & 0.20 $\pm$ 0.05 & 0.46 $\pm$ 0.01 & 0.45 $\pm$ 0.03 \\

% \addlinespace
\midrule
% \addlinespace

\textbf{PPOaug}
    & 0.259 $\pm$ 0.010 & 0.0212 & 2378 $\pm$ 101 & \textbf{0.0046}  \\
    % & $\|\omega\|$ (\unit{rad / s}) & 0.48 $\pm$ 0.01 & 0.42 $\pm$ 0.01 & 0.48 $\pm$ 0.01 & 0.74 $\pm$ 0.04 & 0.62 $\pm$ 0.01 \\

% \addlinespace
\midrule
% \addlinespace

\textbf{\gls{PPOsymm}}
    & \textbf{0.254} $\pm$ 0.014 & \textbf{0.0207} & \textbf{2026} $\pm$ 187 & 0.0172  \\
    % & $\|\omega\|$ (\unit{rad / s}) & 0.81 $\pm$ 0.10 & 0.82 $\pm$ 0.14 & 0.88 $\pm$ 0.08 & 1.06 $\pm$ 0.18 & 0.82 $\pm$ 0.01 \\

% \addlinespace
\bottomrule
% \addlinespace

\end{tabular}

\caption{
    \label{tab:standdance-performance-metrics} Comparison of command tracking error, Cost of Transport and their symmetry index on stand turning tasks for three \gls{PPO} variants. \gls{PPOsymm} demonstrates less error and energy consumption, indicating a more optimal policy.
    \vspace*{-0.4cm}
}

\end{table}

\noindent
% In this bipedal locomotion task, we explore how incorporating symmetry may help in the case of motion symmetry. 
% In this , we explore how symmetry aids motio. 
Besides loco-manipulation, we examine how symmetry helps if the task involves bipedal locomotion only. 
Qualitatively, the \gls{PPO} policy repeatedly develops a staggered gait with one foot positioned ahead of the other to maintain balance. This gait results in significant jittering and hinders symmetric behaviors between left and right turns, as shown in Fig. \ref{fig:stand-dance}. Furthermore, even though the reward function explicitly encourages symmetry, \gls{PPO} fails to transfer on hardware unless the reward terms are carefully tuned to generate a relatively symmetric gait. Conversely, symmetry-incorporated policies exhibit both symmetric gaits and more importantly symmetric turning trajectories in the task level (Fig. \ref{fig:stand-dance}). 

To measure the optimality of the policies, we gauge the error between the robot's commanded and actual headings. 
Additionally, we include the \gls{cot} to evaluate the energy efficiency of the controller. Here, we calculate \gls{cot} as $\frac{\sum_t\sum_{\text{12 actuators}}[\tau \dot{\theta}]^{+}}{\sum_t |v|}$, where $\tau$ is joint torque, $\dot{\theta}$ is joint angular velocity, and $v$
is the horizontal velocity of the robot base at each simulation timestep. 
We observe that \gls{PPOsymm} achieves both lower tracking error 
and \gls{cot} compared to other variants. This indicates the relative optimality induced by the hard equivariance constraint. In comparison, \gls{PPOaug}
achieves lower tracking error and error SI but higher \gls{cot} than PPO, indicating that the behavior is similarly symmetric but less optimal compared to \gls{PPOsymm}. 

\subsection{Slope Walking Task}
\begin{figure}[t]
  \centering
  \def\smallgap{5pt}
  \def\height{0.13\textwidth}
  \begin{tikzpicture}[inner sep=0pt]
      \node (ppo) {\includegraphics[height=0.133\textwidth]{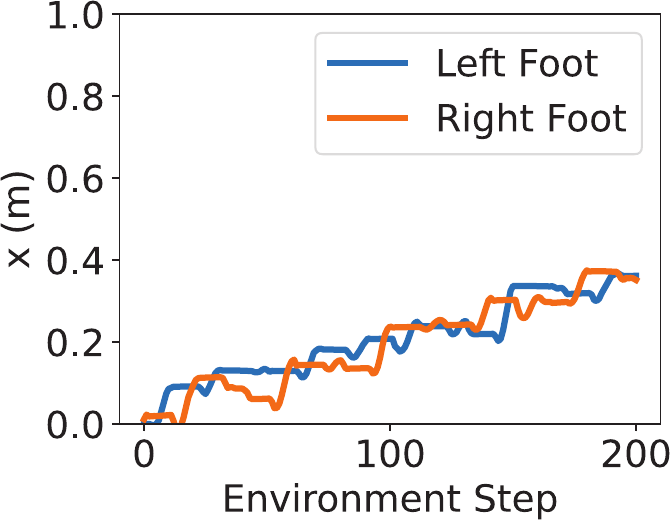}};
      \node[right=\smallgap of ppo] (ppoaug) {\includegraphics[height=\height]{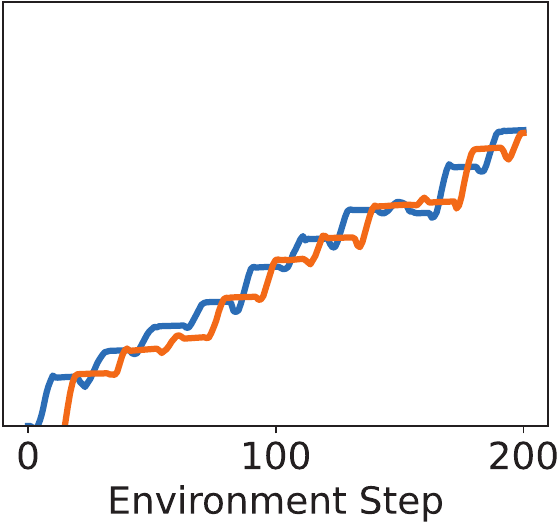}};
      \node[right=\smallgap of ppoaug] (ppoeq) {\includegraphics[height=\height]{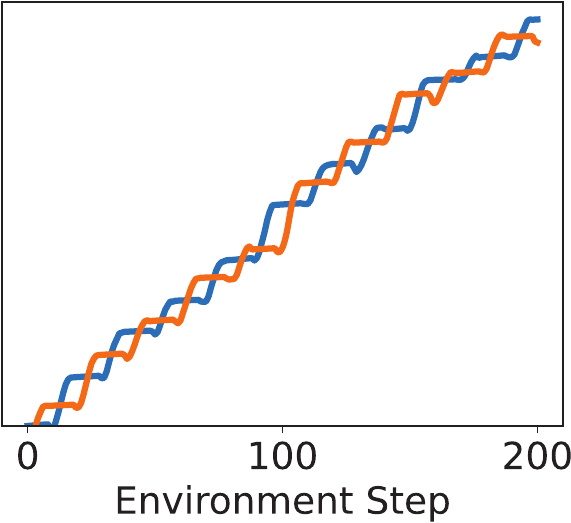}};
      \node[black, anchor=south east, yshift=12pt, inner sep=2pt] at (ppo.south east) {\footnotesize (a)PPO};
      \node[black, anchor=south east, yshift=12pt, inner sep=2pt] at (ppoaug.south east) {\footnotesize (b)PPOaug};
      \node[black, anchor=south east, yshift=12pt, inner sep=2pt] at (ppoeq.south east) {\footnotesize (c)PPOeqic};
  \end{tikzpicture}
\caption{Plots of the feet positions in the desired walk direction. We observe that vanilla \gls{PPO} learns an unstable step pattern with backward steps and foot slipping, resulting in $50\%$ slower walking speed. \gls{PPOaug} improves drastically but asymmetric patterns such as foot dragging still exists. \gls{PPOsymm} presents the most symmetric interweaving gait pattern and walks at the desired speed.}
\label{fig:walk-slope} 
\end{figure}
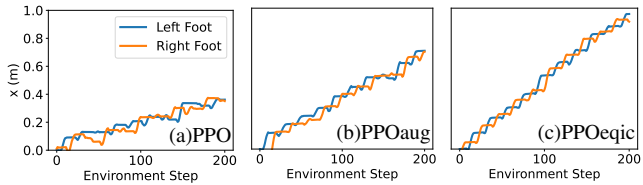

\noindent
In the challenging task of slope walking, we see significant differences between the three \gls{PPO} variants. The \gls{PPO} policy learns a more natural gait compared to other tasks. However, it still shows significant issues with learning a symmetric foot placement, resulting in backward steps and frequent balance loss. This is evident from the plot of the feet positions in the direction of desired motion, shown in Fig. \ref{fig:walk-slope}(a), where \gls{PPO} shows undesired behaviors including stepping in the same place, stepping backwards, or slipping on the ground. Additionally, the locomotion gaits are devoid of periodic stepping in some instances. This unregulated gait pattern leads to significant decrement in velocity tracking performance, where the policy walks only half as far as \gls{PPOsymm} in the same timeframe. 

The \gls{PPOaug} policy improves upon PPO's limitations, exhibiting better alternation between the leading foot and more regulated foot placement. However, as shown in Fig. \ref{fig:walk-slope}(b) variations in step size and occasional staggering of one leg persist, indicating room for improvement. 

\gls{PPOsymm} presents the most stable gait. Shown in Fig. \ref{fig:walk-slope}(c), it maintains consistent foot exchange, regulated contact sequences, and similar step sizes, even on a $11.3^{\circ}$ incline. This demonstrates \gls{PPOsymm}'s effectiveness in learning challenging tasks, highlighting its optimal performance through the hard equivariance constraints compared to other methods. Quantitatively, it is able to walk at the desired velocity of 0.25 $m/s$, $20\%$ more accurately than \gls{PPOaug}. 

\subsection{Real-world Experiments} \label{sec:real-world}
\noindent
In this section, we explore the sim-to-real transfer of these methods through real-world experiments on two selected tasks, thereby evaluating their efficacy beyond simulation.

\begin{figure}[t]
   \centering
   \def\smallgap{0pt}
   \begin{tikzpicture}[inner sep=0pt]
   \node (pushdoorright) {\includegraphics[width=0.95\linewidth]{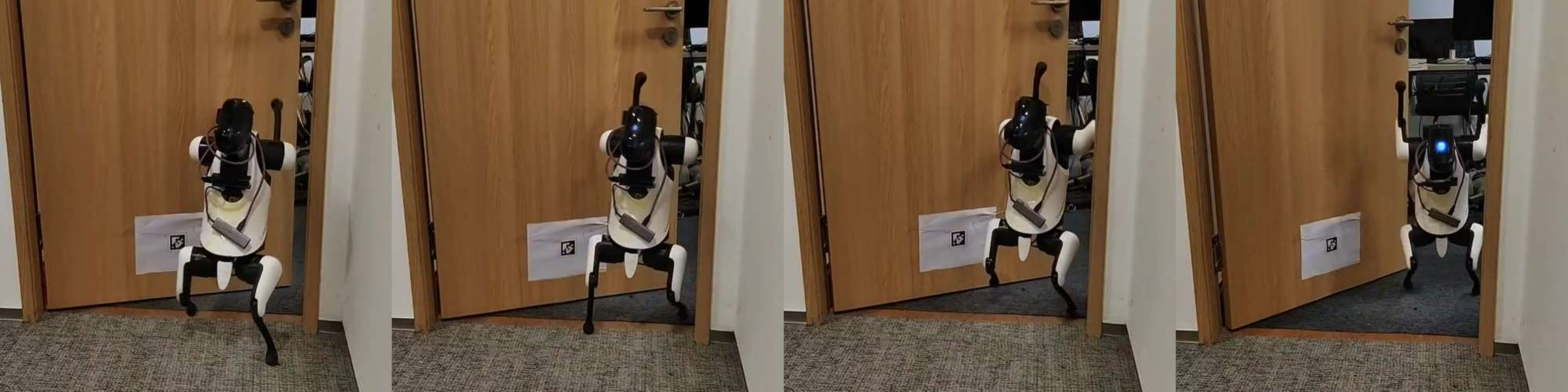}};
   \node[below=\smallgap of pushdoorright] (pushdoorleft) {\includegraphics[width=0.95\linewidth]{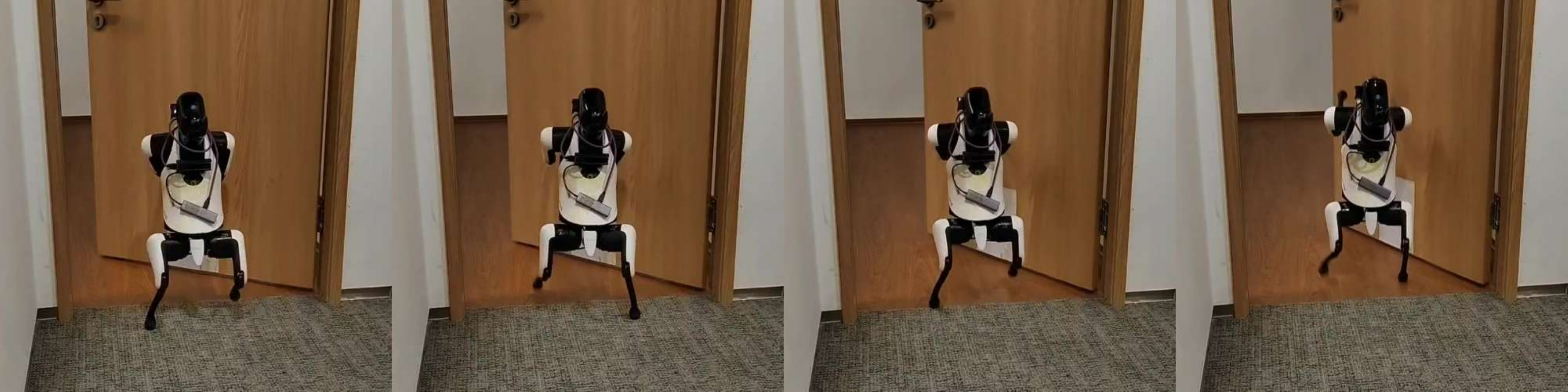}};
   \node[below=\smallgap of pushdoorleft] (standdanceleft) {\includegraphics[width=0.95\linewidth]{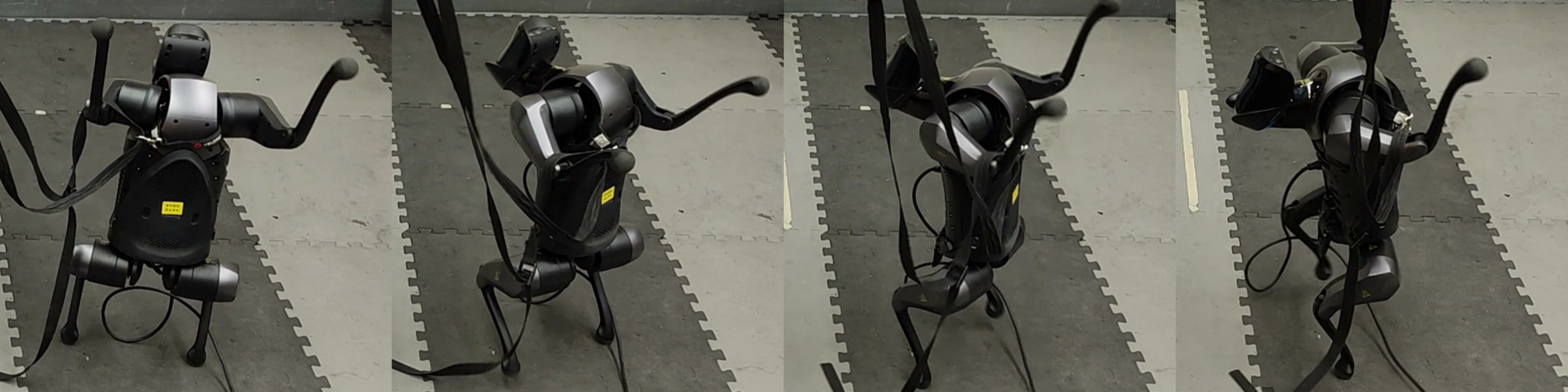}};
   \node[below=\smallgap of standdanceleft] (standanceright) {\includegraphics[width=0.95\linewidth]{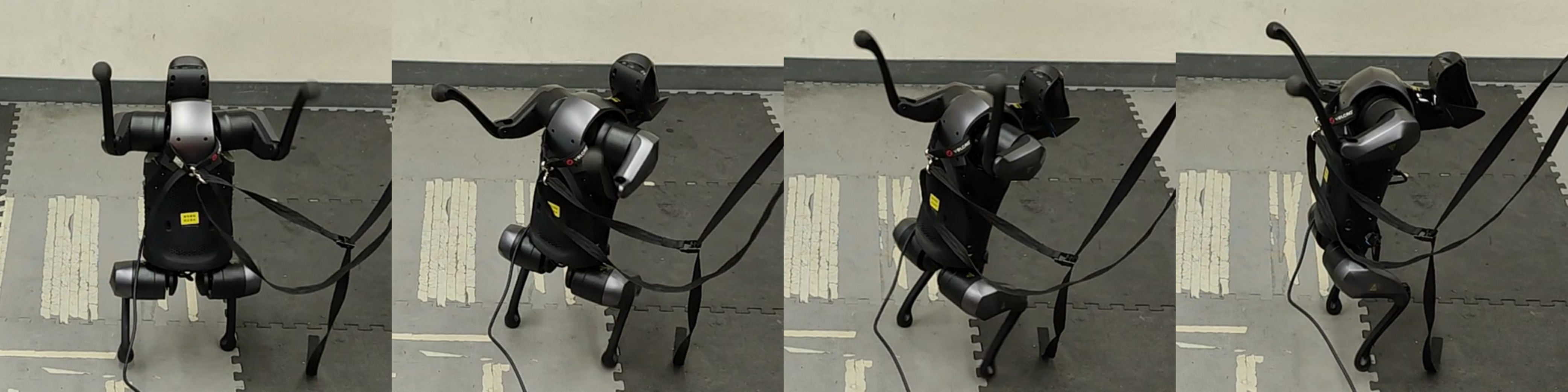}};
   \node[white, anchor=south west, inner sep=2pt] at (pushdoorright.south west) {(a) Left};
   \node[white, anchor=south west, inner sep=2pt] at (pushdoorleft.south west) {(b) Right};
   \node[white, anchor=south west, inner sep=2pt] at (standdanceleft.south west) {(c) Left};
   \node[white, anchor=south west, inner sep=2pt] at (standanceright.south west) {(d) Right};
   \end{tikzpicture}
\caption{
    Snapshots of equivariant policies deployed zero-shot to perform real-world tasks of door pushing (a,b) and stand turning (c,d). For task-level symmetry (e.g. door pushing), \gls{PPOsymm} trained with only right-handed door can be deployed on both left and right-handed doors with symmetric gait patterns. For motion symmetry, \gls{PPOaug} is more robust against slightly asymmetric dynamics that exists on actual hardware. 
    \vspace*{-0.5cm}
}
\label{fig:real} 
\end{figure}
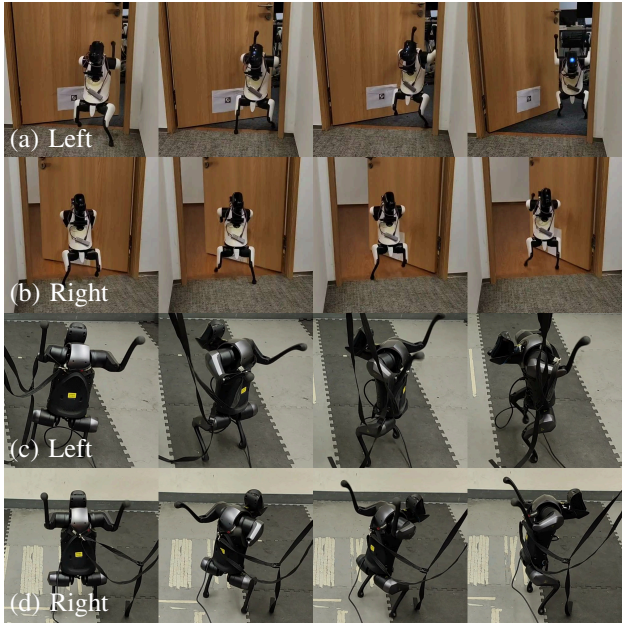

\subsubsection{Door Pushing Task}
% Although vanilla \gls{PPO} policy demonstrates high success rates in the simulation, its performance in real-world scenarios remains limited. The highly asymmetric gaits learned by \gls{PPO} cannot keep the robot from falling over, potentially because it overfits to the simulated contact dynamics, which inevitably differs from the real world, due to simulation inaccuracy. Contrastingly, \gls{PPOsymm} showcases better robustness. Even though trained exclusively on right-handed doors, it is able to locomote steadily and open both left and right-handed doors in real-world settings. Unlike \gls{PPOsymm}, \gls{PPOaug} tends to often reply on just one front leg and demonstrate limited strength in pushing doors. 
% Showing better real-world performance, we show the importance of equivariance constraints in achieving more optimal behaviors. 
Although vanilla \gls{PPO} achieves high success rates in simulation, its real-world performance is limited. The asymmetric gaits learned by \gls{PPO} often cause the robot to fall, likely due to overfitting to simulated contact dynamics that differ from the real world. In contrast, \gls{PPOsymm} is more robust: even though trained exclusively on right-handed doors, it is able to locomote steadily and open both left and right-handed doors. On the other hand, \gls{PPOaug} tends to rely on one front leg, showing limited strength in pushing doors. This highlights the importance of equivariance constraints for optimal real-world behavior.

\subsubsection{Stand Turning Task}
Similar to the door pushing task, the vanilla \gls{PPO} policy is able to locomote without falling over in simulation but fails in real-world contexts. In comparison, both \gls{PPOaug} and \gls{PPOsymm} policies demonstrate good zero-shot transferability, executing 90-degree turns to both sides with high consistency. Notably, the \gls{PPOaug} policy exhibits significantly higher robustness and stability, successfully completing the task in 9 out of 10 trials. This again highlights the substantial enhancement in sim-to-real transfer by incorporating symmetry into the learning process. 

\subsubsection{Discussion}
The intricacies of real-world applications often reveal that environments are not perfectly symmetric MDPs. Discrepancies such as uneven robot mass distribution and variance in actuator dynamics introduce inherent asymmetries in the robot's dynamics. 
% Our findings indicate that in loco-manipulation tasks, where task-space symmetry dominates, the asymmetry introduced by the robot's intrinsic properties is negligible compared to the overarching symmetry within the task space. 
% In this case, policies with hard symmetry constraints, (e.g., \gls{PPOsymm} policy), excel in real-world task as well.
Our findings suggest that in loco-manipulation tasks, where task-space symmetry dominates, the robot's intrinsic asymmetries are negligible for the overall MDP, allowing hard symmetry constraints (e.g., \gls{PPOsymm}) to excel in real-world scenarios. 
However, in bipedal walking tasks which consist of almost entirely intrinsic symmetry, we observed that \gls{PPOsymm} is more vulnerable to distribution shifts compared to methods without hard constraints (e.g., \gls{PPOaug}). By allowing natural adaptation to the robot's asymmetries during training, \gls{PPOaug} demonstrates enhanced robustness against such imperfect symmetry. This aligns with prior work~\cite{mittal2024symmetry}, suggesting that the choice between \gls{PPOaug} and \gls{PPOsymm} should be tailored to the specific nuances of the problem.

%% CONCLUSIONS
\section{Conclusion}
\noindent
In this work, we investigate the benefits of leveraging symmetry in model-free RL for legged locomotion. We compare the performance of incorporating data augmentation and hard equivariance constraints across four challenging bipedal locomotion and loco-manipulation tasks against a vanilla \gls{PPO} baseline. We find that imposing hard symmetry constraints on the learned policy and value functions leads to better performance than other methods. Compared to vanilla PPO, equivariant policies learn notably more steady and symmetric gait patterns, eventually leading to better task-space symmetry. More importantly, equivariant policies trained on single symmetry mode are directly generalizable to other modes.
When applied to real-world scenarios, symmetry-incorporated policies demonstrate significantly better robustness than unconstrained policy. Furthermore, \gls{PPOaug} copes with slight asymmetry in robot's own dynamics, while \gls{PPOsymm} demonstrates better performance on task-space symmetry. 

% Limitation
% Though not encountered in this work, hard constraints can sometimes overly limit the exploration, needing to be addressed in future works. 

We hope this work can aspire the exploration of leveraging symmetry constraints on robots with larger symmetry groups than the reflection group concerned in this work. As the number of symmetry modalities increases, the symmetry constraints are expected to play a more crucial role in guiding the exploration of model-free RL over the increasingly complex state, action spaces. 
In addition, equivariant policies demonstrate promising potential for even larger performance gains over vanilla PPO, highlighting improvements as large as $26\%$ in some seeds. Future efforts could be on stabilizing training to consolidate this enhancement and develop better symmetry-incorporated RL algorithms.

\section*{Acknowledgements} 
\noindent
X. H., Z. L., Q. L., and K. S. acknowledge financial support from The AI Institute, InnoHK of the Government of the Hong Kong Special Administrative Region via the Hong Kong Centre for Logistics Robotics. G. T., M. P., and C. S. acknowledge financial support from PNRR MUR Project PE000013 "Future Artificial Intelligence Research (hereafter FAIR)", funded by the European Union – NextGenerationEU. The authors thank Prof. Xue Bin Peng for insightful discussions on this work. The authors also thank Xiaomi Inc. for providing CyberDog 2 for experiments.

%% BIBLIOGRAPHY
{
\bibliographystyle{IEEEtran}
\bibliography{bib/bibliography}
}

%%% ACRONYMS

{\small
    \let\oldsection\section % Save the old section command
    % Redefine the section command to use \normalsize and make it unnumbered
    \renewcommand\section[2]{\oldsection*{\normalsize #2}}
    \printglossary[title=Special Terms, style=index, nogroupskip]
    \let\section\oldsection % Restore the original section command
}

% \begin{acronym}
% \acro{HP}{high-pass}
% \acro{LP}{low-pass}
% \end{acronym}

\end{document}